%% file: main.tex
\documentclass[10pt,twocolumn,letterpaper, pagenumbers]{article}
\usepackage{arxiv}              %

\input{author-kit/preamble}
\definecolor{arxivblue}{rgb}{0.21,0.49,0.74}
\usepackage[pagebackref,breaklinks,colorlinks,citecolor=arxivblue]{hyperref}
\usepackage{amsmath}
\usepackage[table]{xcolor} %
\usepackage{makecell}

\newcommand\blfootnote[1]{%
  \begingroup
  \renewcommand\thefootnote{}\footnote{#1}%
  \addtocounter{footnote}{-1}%
  \endgroup
}

\title{NeuralFur: Animal Fur Reconstruction From Multi-View Images}

\author{
Vanessa Sklyarova$^{1,2*}$ \and 
Berna Kabadayi$^{1,4*}$ \and
Anastasios Yiannakidis$^{1}$ \and 
Giorgio Becherini$^{1}$ \and 
Michael J. Black$^{1}$ \and
Justus Thies$^{1,3}$ \and 
\vspace{0.1cm}\\
$^1$Max Planck Institute for Intelligent Systems \ \ $^2$ETH Zürich \\ $^3$Technical University of Darmstadt \ \ $^4$University of Tübingen
}

\begin{document}

\twocolumn[{
\renewcommand\twocolumn[1][]{#1}%
\maketitle

\begin{center}
    \centering
    \captionsetup{type=figure}
    \vspace{-0.5cm}
    \includegraphics[width=\textwidth]{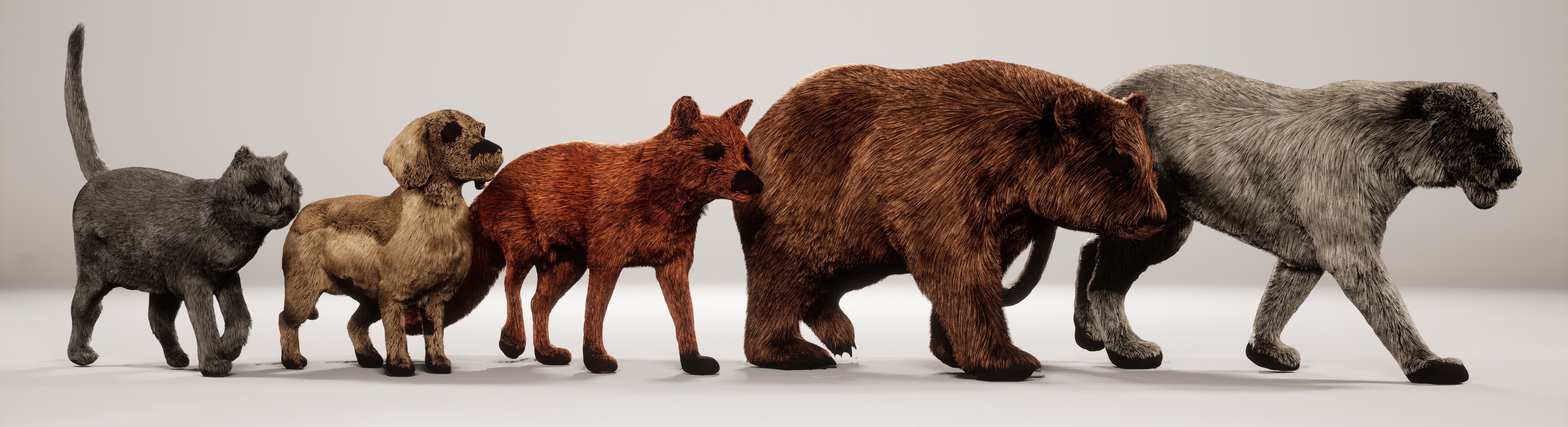}
    \vspace{-0.5cm}
    \caption{
        From multi-view images, \textit{NeuralFur} reconstructs detailed geometries of animals with a mesh-based body and strand-based fur.
        The reconstructions can be integrated in computer graphics frameworks for simulation and rendering with artist-defined colors.
    }
    \label{fig:teaser}
\end{center}
}]

\blfootnote{$^*$ Equal contribution}

\input{author-kit/sec/abstract}

\input{author-kit/sec/intro}

\input{author-kit/sec/related_works}
\input{author-kit/sec/method}

\input{author-kit/sec/experiments}

\input{author-kit/sec/conclusion}

\input{author-kit/sec/acknowledgement}

{
    \small
    \bibliographystyle{ieeenat_fullname}
    \bibliography{main}
}

\clearpage

\maketitlesupplementary
\input{author-kit/supp_sec/experiments}

\clearpage

\end{document}

%% file: author-kit/preamble.tex
\usepackage[dvipsnames]{xcolor}
\usepackage{comment}

\def\S{\mathbf{S}} %
\def\p{\mathbf{p}} %
\def\zgeom{\mathbf{z}} %
\def\x{\mathbf{x}} %
\def\E{\mathcal{E}} %
\def\G{\mathcal{G}} %

%% file: author-kit/sec/abstract.tex
\begin{abstract}
Reconstructing realistic animal fur geometry from images is a challenging task due to the fine-scale details, self-occlusion, and view-dependent appearance of fur.
In contrast to human hairstyle reconstruction, there are also no datasets that can be leveraged to learn a fur prior for different animals.
In this work, we present a first multi-view-based method for high-fidelity 3D fur modeling of animals using a strand-based representation, leveraging the general knowledge of a vision language model.
Given multi-view RGB images, we first reconstruct a coarse surface geometry using traditional multi-view stereo techniques.
We then use a vision language model (VLM) system to retrieve information about the realistic length structure of the fur for each part of the body.
We use this knowledge to construct the animal’s furless geometry and grow strands atop it. 
The fur reconstruction is supervised with both geometric and photometric losses computed from multi-view images.
To mitigate orientation ambiguities stemming from the Gabor filters that are applied to the input images, we additionally utilize the VLM to guide the strands' growth direction and their relation to the gravity vector that we incorporate as a loss.
With this new schema of using a VLM to guide 3D reconstruction from multi-view inputs, we show generalization across a variety of animals with different fur types.  
For additional results and code, please refer to
\url{https://neuralfur.is.tue.mpg.de}.

\end{abstract}

%% file: author-kit/sec/intro.tex
\section{Introduction}
\label{sec:intro}

Animals frequently appear as characters in animated films.
The realism of these characters relies on strand-based models of the fur, which are ideal for physics simulation and provide high realism.
Unfortunately, the manual creation of fur grooms for animals is a time-consuming process for experienced artists~\cite{seymour2020weta}.
In this work, we study the automatic generation of fur groom geometry, which can be directly applied to background characters or used as an initialization for hero-level assets.
While the automated creation of strand-based human hair grooms from text \cite{sklyarova2024text}, scans \cite{lazuardi2025geomhairreconstructionhairstrands, shen2023CT2Hair}, images~\cite{Sklyarova2025im2haircut, zhou2024groomcap, scan2hair, difflocks2025, Takimoto_2024_CVPR, Rosu2022NeuralSL, Kuang2022DeepMVSHairDH} or videos \cite{sklyarova2023neural_haircut, zakharov2024gaussianhaircut, monohair} has been widely studied, there is no equivalent work on animal fur.

Unlike human hair, animal fur tends to cover the entire body, varies across different animals, and differs in length, density, and orientation even within the same animal. 
Another difficulty is that there are no large datasets of animals with fur that can be leveraged to learn
a prior for the inner fur structure and the global fur style to constrain optimization-based reconstruction from a set of images.
Thus, we have to ask how can we get information about how hair varies across animals and over the body of an animal?
Our key idea is to leverage both given multi-view images, where we can extract orientations and silhouettes, and an vision language model (VLM) that is used to resolve ambiguities from the image signal and to provide general knowledge about fur of the specific animal.
This idea leads to our method called \textit{NeuralFur}, which enables automatic reconstruction of a variety of 3D animals with detailed strand-based fur from given multi-view images.
\textit{NeuralFur} is the first method that automatically reconstructs animal fur that can be imported into a standard computer graphics frameworks for simulation and rendering.

While recent methods~\cite{biggs2020wldo, niewiadomski2024generativezoo, lyu2025animer, sabathier2024animal} show interesting results for reconstructing 3D animals from images and videos using the mesh-based, parametric SMAL~\cite{Zuffi:CVPR:2017} model, they are lacking geometrical details like fur.
In contrast, we aim for a realistic digital replica of the geometry of an animal with the following properties: 
(1) The fur is represented with strands, which are compatible with existing rendering and simulation pipelines.
(2) The underlying geometry reflects the true geometry of an animal rather than a coarse approximation of it.
(3) Animal body parts are taken into account during 3D reconstruction as different parts of the animals differ in strand length. %
To this end, we propose a method that leverages recent advances in human hair reconstruction and general knowledge from vision language models to reconstruct detailed 3D animal models with fur.
Specifically, we build our method based on the state-of-the-art strand-based human-hair reconstruction technique Gaussian Haircut~\cite{zakharov2024gaussianhaircut}, where hair is represented with connected Gaussian primitives.
Based on multi-view images, we reconstruct a fine-grained geometry using NeuS~\cite{Wang2021NeuSLN}.
We use a vision language model to reason about fur thickness, growth direction, and length.
This information is used to refine the NeuS-based reconstruction, and to guide the 3DGS-based fur strand reconstruction.

With this scheme, we address the following key challenges:
(i) the NeuS reconstruction results in a geometry that includes the fur.
To obtain the actual geometry of the underlying body requires considering the thickness / volume of the fur which can vary largely depending on the body part region (e.g., fur is typically very short in the face region, whereas there could be very long hair in other regions like at the mane of a lion).
To localize the body parts of an animal, we fit the SMAL~\cite{Zuffi:CVPR:2017} model to the NeuS reconstruction and transfer general body part labels from SMAL to the NeuS reconstruction.
Given an image of an animal and the body part labels, we leverage a VLM to retrieve information about the local fur such that we can remove the fur from the geometry (defurring).
This step allows us to model the actual body of the animal, where we can place the roots for our fur strands.
(ii) animal fur grows in different directions within different animal body parts.
As orientation maps extracted from the input images based on Gabor-filters are undirected, we leverage a vision language model to extract a general guiding signal for the growing direction to resolve the ambiguity.
To regularize fur-strand geometries during optimization, we employ a strand-level prior learned from human hair, as no fur-specific datasets are available. Although not ideal, this approach proves sufficient for our purposes.

As illustrated in Figure \ref{fig:teaser}, our method, for the first time, is effective in reconstructing entire geometries of animals with fur from multi-view images.
\textit{NeuralFur} generalizes across different animal types, and takes their region-dependent fur properties into account.
The fur is compatible with standard computer graphics pipelines, which enables downstream applications with physics-based rendering and simulation.
Note that we do not target appearance reconstruction; colors and shader attributes can be assigned to our reconstructed fur strand geometries in a downstream rendering engine, as shown in Figure \ref{fig:teaser}. 
Extensive qualitative comparisons with prior animal- and hair-reconstruction methods demonstrate that, unlike previous approaches, our method more accurately reconstructs detailed animal fur geometry.

\medskip\noindent
In summary, we present the first strand-based fur reconstruction method for animals from multi-view images.
It is enabled by the following contributions:
\begin{itemize}
    \item the hybrid usage of images and general knowledge from a VLM model, where the external knowledge is used to resolve ambiguities from the input to estimate the underlying animal body and to guide the growing direction and fur length.
    A key aspect is the localization of the retrieved VLM information to specific regions of the animal.
    \item an implicit fur representation network based on a multi-layer perceptron (MLP), which enables the prediction of spatially varying fur strands.
\end{itemize}

%% file: author-kit/sec/related_works.tex
\input{author-kit/figures/method_fig}
\section{Related works}
\label{sec:related_works}

We propose a method for reconstructing detailed 3D animal geometry with strand-based fur from multi-view images, guided by a vision-language model.
Our approach unifies classical mesh-based animal reconstruction with strand-based human hair techniques.
\paragraph{Animal reconstruction.}
Creating a 3D digital replica of animals presents numerous challenges.
These include recovering the pose~\cite{pereira2022sleap, yang2021lasr, yu2021ap}, shape~\cite{xu2023animal3d, biggs2018creatures}, and appearance of an animal from images~\cite{Ruegg_2023_CVPR, wu2023magicpony} or videos~\cite{NEURIPS2021_a11f9e53, Yang_2021_CVPR, casa, yang2021viser, yang2022banmo, sabathier2024animal}.
Despite recent progress, relatively few works address 3D animal reconstruction and appearance~\cite{Yan2017efficient, Yan2015Physically-accurate, andersen2016hybrid, artemis}, primarily due to the scarcity of 3D animal datasets~\cite{xu2023animal3d, yu2021ap} and the inherent difficulty of capturing static and dynamic animals~\cite{artemis}.
Early work focuses on category-specific reconstruction, such as dogs~\cite{biggs2020wldo, kearney2020rgbd, Ruegg_2023_bite}, horses~\cite{zuffi2019three, li2021hsmal}, and birds~\cite{kanazawa2018learning}.
Cashman and Fitzgibbon~\cite{cashman2012shape} learn a parametric dolphin model from images, LASSIE~\cite{yao2022lassie} learns articulated animal shape via part discovery. Kanazawa et al.~\cite{kanazawa2016learning} deform template meshes for cats from correspondences, and \cite{kearney2020rgbd} predict canine pose. 
Zuffi et al.~\cite{Zuffi:CVPR:2017} introduced SMAL, a parametric model for quadruped animals.
GenZoo~\cite{niewiadomski2024generativezoo} uses SMAL~\cite{Zuffi:CVPR:2017} and trains a regressor on synthetic data to estimate pose and shape from a single image.
AniMer~\cite{lyu2025animer} estimates animal pose and shape using a family-aware transformer, enhancing the reconstruction accuracy of diverse quadrupedal families.
AWOL~\cite{zuffi2024awol} generates animals and trees using language guidance. RAW~\cite{kulits2025reconstructing} reconstructs animals and their surroundings.
Both model-based approaches~\cite{xu2023animal3d, niewiadomski2024generativezoo, lyu2025animer} (i.e., methods that rely on parametric models for animal representation) and model-free approaches~\cite{yang2023ppr, wang2021birds}  typically rely on coarse geometry to represent animals, providing a low-dimensional representation. However, animals often exhibit complex geometries, including fur and fine surface details. In this work we propose a \textit{layered representation} consisting of a mesh for body with a strand-based fur layer that integrates seamlessly with classical graphics pipelines for simulation, rendering, and editing

\paragraph{Strand-based reconstruction.}
Strands are the standard representation for high-fidelity 3D hair modeling in research~\cite{Yuksel2009HairM,piuze2011generalized,shen2023CT2Hair} and production~\cite{chiang2016practical,fascione2018path, scan2hair}.
This parametrization offers advantages over volumes or meshes in terms of physics simulation~\cite{fei2017multi,daviet2023interactive,hsu2023sagfree} and geometric control~\cite{xing2019hairbrush,shen2020deepsketchhair,sklyarova2024text,zhou2023groomgen}.
However, manual modeling remains labor-intensive due to the high strand count and geometric complexity.
To automate this process, image-based methods estimate strand orientations from photographs~\cite{Paris2004CaptureOH} and optimize 3D strands to align with 2D projections~\cite{paris2008hair,luo2012multi,chai2015high,Chai2016AutoHairFA,Nam2019StrandAccurateMH, Sklyarova2025im2haircut}.
Yet, these approaches struggle with occlusions and typically model strands as polylines without thickness, separating geometry and appearance~\cite{Rosu2022NeuralSL,sklyarova2023neural_haircut, monohair, Takimoto_2024_CVPR}.
Recent works incorporate learned priors~\cite{sklyarova2023neural_haircut, monohair, Rosu2022NeuralSL} or anisotropic 3D Gaussians for richer geometry and appearance representation~\cite{Luo2024GaussianHairHM,zakharov2024gaussianhaircut, zhou2024groomcap}.
While sharing detail-preservation challenges with hair reconstruction methods, strand-based fur capture must additionally address regional fur length modeling, part segmentation, and furless geometry estimation.

\paragraph{Large vision-language models.}
Vision–language models (VLMs) combine visual perception with language understanding, enabling open-ended reasoning about images.
Large-scale pretraining approaches, such as CLIP~\cite{Radford2021clip}, show strong generalization across diverse visual tasks, as do more recent models like BLIP~\cite{li2022blip}, LLaVA~\cite{liu2023llava, liu2023improvedllava}, and Flamingo~\cite{flamingo}.
More recently, commercial systems such as GPT-5, Gemini, and Claude have demonstrated fast and efficient multi-modal reasoning, including the ability to process multiple images simultaneously.
In this work, we use ChatGPT~\cite{openai2025gpt5} to support accurate modeling of realistic animal fur geometry.

%% file: author-kit/figures/method_fig.tex
\begin{figure*}[t]
    \centering
    \includegraphics[trim=0cm 65.7cm 0cm 0cm, clip,width=\linewidth]{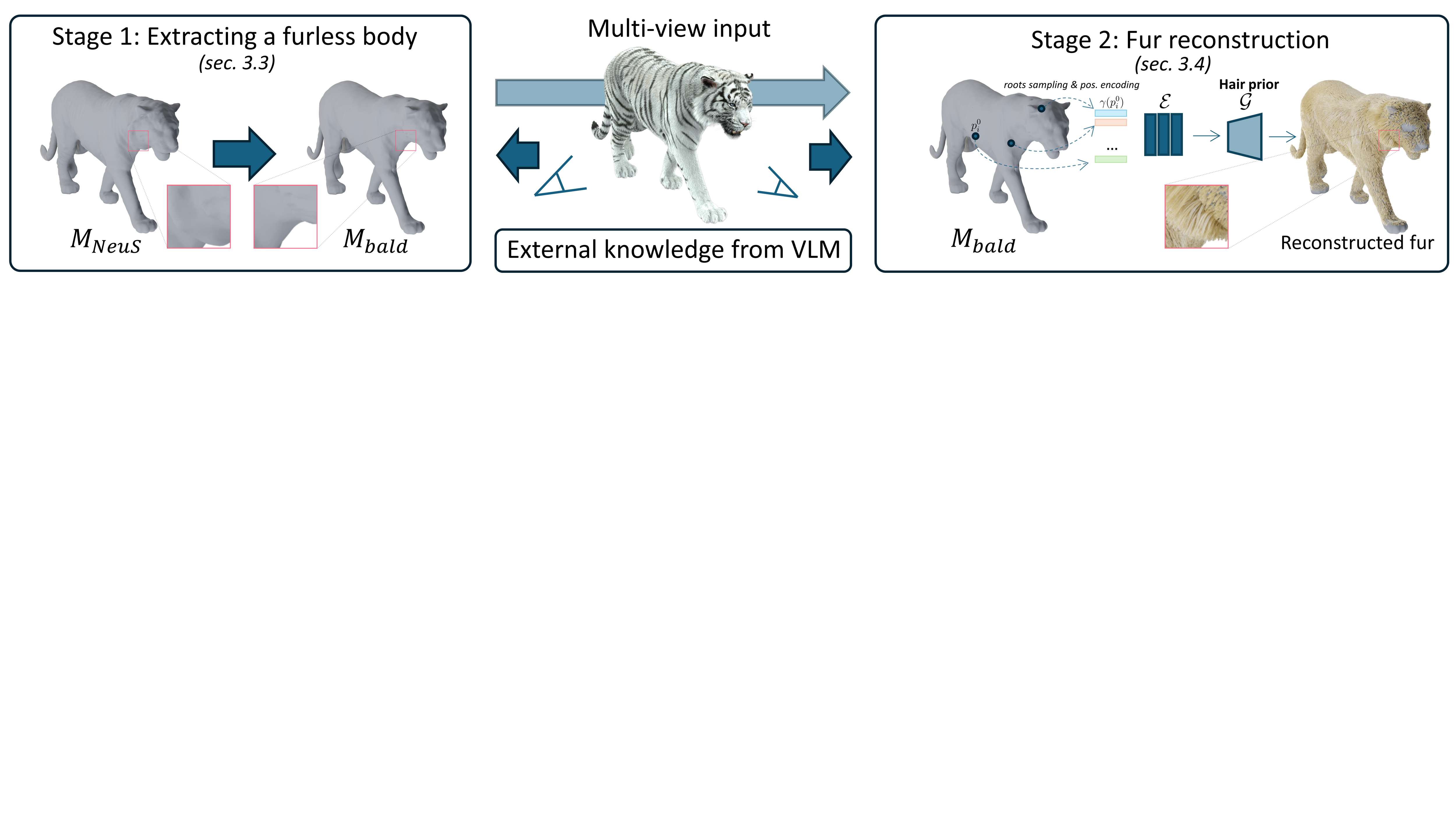}

    \caption{
    Our method, \textit{NeuralFur}, consists of two stages: (i) extracting a furless mesh geometry by intelligently shrinking the full mesh reconstructed from multi-view images, and (ii) reconstructing strand-based fur by initializing roots from the furless mesh.
    For both stages, external knowledge from a VLM is leveraged. Based on the depicted animal, the VLM provides information about fur thickness, length, and orientation.
    This guidance is then used to train a neural fur strand representation (MLP), which can be queried at any mesh surface location to generate fur strands suitable for rendering.
    }
    \label{fig:method-details}
\end{figure*}

%% file: author-kit/sec/method.tex
\section{Method}
\label{sec:method}

Our method consists of two stages: (i) extracting a furless mesh geometry of the animal from multi-view images and (ii)  strand-based fur reconstruction, see \Cref{fig:method-details}.
We initialize our method with NeuS~\cite{Wang2021NeuSLN} which results in a mesh that models the entire animal, including fur.
To decompose this mesh into a body mesh and a fur layer, we take advantage of the external knowledge of a VLM that we localize to the different regions of the animal (\Cref{sec:localization}).
Specifically, we extract fur thickness for ``de-furring'' (\Cref{sec:defurring}) as well as fur length information for fur growing from this model (\Cref{sec:reconstruction}).

\input{author-kit/figures/part_annotation}

\subsection{Shape initialization}
We reconstruct the coarse geometry of an animal using NeuS~\cite{Wang2021NeuSLN} which leverages a signed distance function (SDF).
Specifically, it uses Multi-layer Perceptrons (MLPs) for color and signed distance prediction by applying a volumetric raymarching algorithm and supervising it with silhouette constraints and an Eikonal penalty function.
From the reconstructed SDF the visible geometry of the animal is extracted with Marching Cubes~\cite{marchingcubes1, marchingcubes2} resulting in our initial mesh  $M_{NeuS}$.

\subsection{Localizing external knowledge}
\label{sec:localization}
Reconstructing fur solely from multi-view images without additional constraints is insufficient as we do not know underlying body parts or fur length.
Therefore, we incorporate external knowledge from a VLM (ChatGPT~\cite{openai2025gpt5}) to annotate body regions with fur lengths in centimeters.
To obtain these annotations, we prompt it with several frontal and side-view images of the animal and request fur length, thickness, and growth direction for each region.
However, the answers from ChatGPT need to be structured and attached to specific regions (resulting in localized knowledge).
To this end, we leverage the parametric animal model SMAL~\cite{Zuffi:CVPR:2017}.
SMAL is a morphable model which can represent a variety of four-legged animals.
It offers a fixed topology which is used for region annotations that is shared across all animals.
We define the following regions on SMAL: \emph{leg (front,rear), paw pads, paws, belly, neck, face, ears, inner ear canal, under tail, eyes, tail, nose tip, body and mane if available}. 

To leverage these annotations, we fit SMAL to the $M_{NeuS}$ and transfer the labels based on nearest neighbors, see \Cref{fig:method-transfer}.
Note that $M_{NeuS}$ offers more geometrical detail than SMAL.
The fitting operates in 3D space using three stages:
(1) in the first stage, we optimize for global translation $\boldsymbol{\gamma}$ and global rotation $\boldsymbol{\theta_0}$ of the SMAL model.
(2) in the second stage, we additionally optimize for shape parameters $\boldsymbol{\beta}$ and joint pose parameters $\boldsymbol{\theta}$.
(3) in the last stage, we optimize for per-vertex deformations, which helps to obtain finer details.
In each optimization step, we sample 20,000 points on our initial mesh reconstruction and, similar to WLDO~\cite{biggs2020wldo}, optimize the respective parameters of the different stages with respect to Chamfer distance, Laplacian smoothness regularization as well as normal consistency.

\subsection{De-furring -- Extracting a furless geometry}
\label{sec:defurring}
As the initial geometry reconstruction $M_{NeuS}$ represents the entire body including fur as a single continuous surface, it cannot directly be used to place roots of fur strands.
Thus, we first have to derive the geometry of the ``bald'' animal body $M_{bald}$.
Unfortunately, no datasets exist that contain paired examples of shaved and unshaved animals.
Therefore, we leverage the localized external knowledge from the VLM.
We query information about the effective thickness of the fur and ``erode'' the geometry correspondingly to retrieve the furless animal by converting $M_{NeuS}$ into an SDF and applying a spatially varying shrinkage that offsets the surface inward:
\begin{equation}
\operatorname{SDF}_{\text{defur}}(x) = \operatorname{SDF}(x) + s(x) ,
\end{equation}
where $s(x)$ is the locally defined shrinkage term based on the thickness value defined for the nearest vertex.
From this SDF, we extract the furless mesh $M_{bald}$ using Marching cubes~\cite{marchingcubes1, marchingcubes2}, see Figure~\ref{fig:bald_geometry}.

\input{author-kit/figures/figure_bald}

\subsection{Fur modeling and reconstruction}
\label{sec:reconstruction}

\paragraph{Neural fur representation.}
Similar to hair reconstruction methods, we parametrize the fur geometry of animals using a strand-based representation with $N$ individual strands.
The $i$-th fur strand is represented as a 3D polyline with $L$ points: $\S_{i}=\{\p_{i}^{l}\}_{l=1}^{L}$.
Based on this polyline representation, we define directions between nearest points as $\mathbf{d}^l_i =  \p^{l+1}_i-\p^l_i$ and normalized directions as $\mathbf{b}^l_i = \mathbf{d}^l_i \big/ \|\mathbf{d}^l_i\|_2$.
As polylines have a high degree of freedom in 3D space, we model them in the latent space $\zgeom \in \mathbb{R}^{64}$ of the synthetic hairstyles prior model $\G$ from NeuralHaircut~\cite{sklyarova2023neural_haircut}.
We find that the strand-level hair prior also serves as a good approximation for fur when scaled appropriately (see below).
The strands are defined in a local coordinate system which is typically derived from a tangent-bitangent-normal (TBN) basis constructed via texture space.
In our scenario, we are operating on the defurred mesh $M_{bald}$ which does not come with a unified texture space as for human heads~\cite{Rosu2022NeuralSL, sklyarova2023neural_haircut, zakharov2024gaussianhaircut}.
However, it is important that the local coordinate systems are smoothly aligned over the entire defurred mesh.
We, therefore, compute a directional face field~\cite{Directional} with further sign consistency resolved using parallel transport~\cite{spivak1979comprehensive} across shared edges.
The bitangent vector is defined as a cross-product between the normal and the tangent.

Instead of modeling a finite set of fur strands rooting at the defurred mesh, we define a continuous function that can be sampled at the surface and outputs the latent code of the strand.
Specifically, we use a multilayer perceptron (MLP) $\E$ that can be queried at a surface point $\p_{i}^{0}$ which defines the strand root on the defurred mesh $M_{bald}$:  
\begin{equation}
    \zgeom_{i} = \E(\gamma(\p_{i}^{0})),
\end{equation}
where $\gamma$ is positional encoding~\cite{Rahaman2018OnTS} and $\zgeom_{i}$ is the latent strand code in the local TBN at the surface point $\p_{i}^{0}$.
To obtain the polyline from the latent fur $\zgeom_{i}$ we use the pretrained hair decoder $\G$ from NeuralHaircut~\cite{sklyarova2023neural_haircut}. 
Unlike prior works on hair modeling, our goal is to achieve precise, centimeter-level control over fur length for each specific body region.
To this end, we normalize the output strand from the prior model and multiply by the length $\ell_j$ that we retrieve from the VLM during optimization:
\begin{equation}
   \p_{i}' = \G(\zgeom_{i}),  ~ \p_{i} = \ell_j \frac{\p'_i} { \|\p'_i\|_2}, ~\p_{i} \in \mathbb{R}^{L~\times~3} .
\end{equation}

\paragraph{Fur reconstruction.}
We optimize the animal fur based on this neural fur representation using geometry and photometric losses obtained through differentiable rasterization of strands using Gaussian Splatting~\cite{kerbl3Dgaussians}. 
Specifically, in each optimization step, we sample a set of strand roots $\{\p_{i}^{0}\}_{i=1}^{N}$ and compute the corresponding polylines by decoding the latent strand code retrieved by the MLP at the root location.
Similar to Gaussian Haircut~\cite{zakharov2024gaussianhaircut}, we attach Gaussian primitives to each strand segment for rendering.
Besides the sampling, every step is differentiable, and we can compute gradients to update the MLP weights with respect to:
\begin{align}
\mathcal{L}_{\mathrm{rec}} =
    & \ \lambda_{\mathrm{sil}} \, \mathcal{L}_{\mathrm{sil}}
    + \lambda_{\mathrm{dir}} \, \mathcal{L}_{\mathrm{dir}}
    + \lambda^{gpt}_{\mathrm{dir}} \, \mathcal{L}^{gpt}_{\mathrm{dir}} \nonumber \\
    & + \lambda_{\mathrm{chm}} \, \mathcal{L}_{\mathrm{chm}}
    + \lambda_{\mathrm{penetr}} \, \mathcal{L}_{\mathrm{penetr}}
    + \lambda_{\mathrm{shape}} \, \mathcal{L}_{\mathrm{shape}} \, .
\end{align}
\vspace{-0.5cm}

\medskip \noindent
\textit{Silhouette loss:} We supervise using a silhouette loss  $\mathcal{L}_{\mathrm{sil}}$, which is $L_{1}$ distance between the rendered and estimated animal silhouette from the input images.

\medskip \noindent
\textit{Orientation loss:} We further supervise orientations using: 
\begin{equation}
    \mathcal{L}_\text{dir} = \sum_p \tau_p \min \{ \text{d} (\beta_p, \hat\beta_p), \text{d} (\beta_p, \hat\beta_p) \pm \pi \} - \log \tau_p,
\end{equation}
where $\text{d}$ denotes the absolute angular difference between the directions, $\hat\beta_p$ denotes the direction in the pixel $p$ obtained using a set of Gabor filters~\cite{Paris2004CaptureOH} and  $\tau_p$ is a rendered confidence factor.

We mask the silhouette and direction losses using a non-baldness mask derived from $M_{NeuS}$.
To construct this mask, we render the textured mesh, assigning a value of 1 to the texture of vertices with a nonzero length and 0 otherwise.
In this way, bald regions (such as around the eyes, nose tip, and paw pads) are discarded from the loss.%

\medskip \noindent
\textit{Orientation consistency loss:}
As orientation maps derived from Gabor filters are undirected, we incorporate an auxiliary loss function that leverages prior knowledge of ChatGPT on normalized hair growing direction for each part $\hat{\mathbf{g}}_i\in \mathbb{R}^{3}$ to enforce directional consistency:
\begin{equation}
\mathcal{L}^{gpt}_{\text{dir}} = \frac{1}{N(L-1)} \sum_{i=1}^N \sum_{l=1}^{L-1}w^{l} \max(0, - \cos {(\mathbf{b}^l_i \cdot \hat{\mathbf{g}}_i)}),
\end{equation}
where $w^{l}=\frac{l-1}{L-1}$ is a linear weight that increases from the root to the tip.

\medskip \noindent
\textit{Chamfer loss:}
For geometry supervision~\cite{sklyarova2023neural_haircut}, we use the one-way
Chamfer distance between the full geometry $M_{NeuS}$ and the learned strands:
\begin{equation}
    \mathcal{L}_{\mathrm{chm}} = \sum_{k=1}^K \big\| \x_k - \p_k \big\|_2^2,
\end{equation}
where $\x_k$ are $K$ random points sampled from $S$ and their nearest points on the strands $\p_k$.

\medskip \noindent
\textit{Curvature consistency loss:}
We employ an additional regularization on the consistency of curvature profiles across the fur.
For strand $i$, we compute bending angles $\mathbf{\theta}_{i}^{l}$ between the nearest strand segments:
\begin{equation}
\mathbf{\theta}_{i}^{l} = \arccos \!\Big( \operatorname{clamp}\big( \mathbf{b}^{l-1}_i \cdot \mathbf{b}^l_i, \,-1, \,1 \big) \Big), 
~l = 2, \dots, L-1 .
\end{equation}
The curvature signature for each strand $\mathbf{\theta}(S_{i})=\{\mathbf{\theta}_{i}^{l}\}_{l=2}^{L-1}$ and mean curvature among $N$ strands is defined as $\bar{\theta} = \frac{1}{N} \sum_{i=1}^{N} \theta(S_i)
$.
With this information, we compute the curvature consistency loss:
\begin{equation}
\mathcal{L}_{\mathrm{shape}} = \frac{1}{N} \sum_{i=1}^{N} \frac{1}{L-2} \|\theta(S_i) - \bar{\theta}\| 
^2 .
\end{equation}

\medskip \noindent
\textit{Penetration loss:}
To prevent interpenetrations into $M_{bald}$, we add an additional penetration loss:
\begin{equation}
\mathcal{L}_{\mathrm{penetr}}= \frac{1}{N}\sum_{i=1}^{N}\max(0, -\operatorname{SDF}_{\text{defur}}(x_{i})) .
\end{equation}

\subsection{Implementation details}
For complete geometry reconstruction, we run NeuS~\cite{Wang2021NeuSLN} for 300,000 iterations.
During training fur reconstruction, we sample 15,000 strands per iteration.
At inference, we reconstruct 500,000 strands, with the ability to scale to larger counts, as we are using an MLP to represent the fur groom.
To model fur in metric space, we first identify the eye keypoints in 3D geometric space and use VLM to estimate the distance between the eye centers in the image.
This measured scale is then used to compute positional offsets, allowing the fur to be grown accurately in centimeters. 
Our method takes around 10 hours on a single A100.

%% file: author-kit/figures/part_annotation.tex
\begin{figure}[t]
    \centering
    \includegraphics[width=\linewidth]{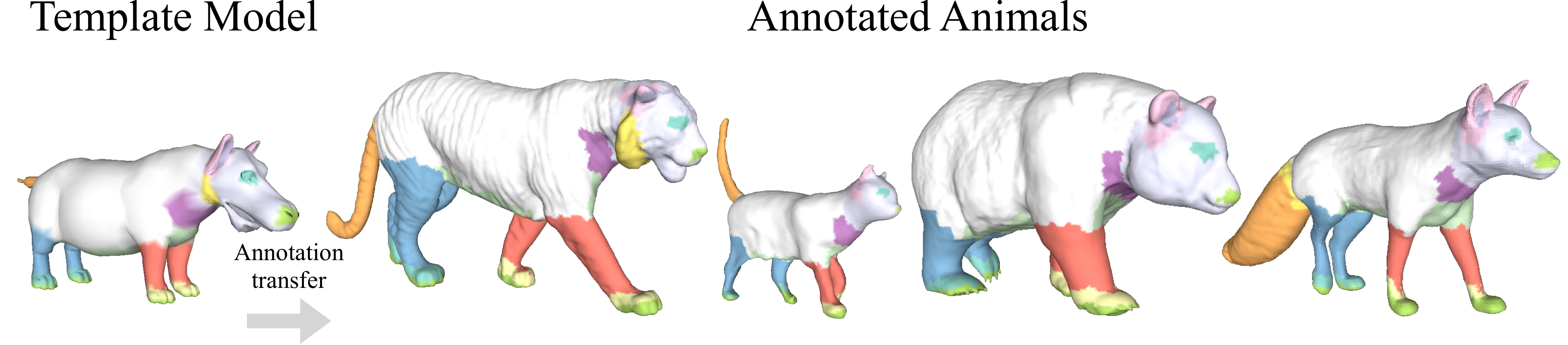}
    \vspace{-0.5cm}
    \caption{\textbf{Annotation.} Automatic part annotation for each animal obtained from the fitted  SMAL~\cite{Zuffi:CVPR:2017} model.}
    \label{fig:method-transfer}
\end{figure}

%% file: author-kit/figures/figure_bald.tex
\begin{figure}[t]
    \centering
    \includegraphics[width=\linewidth]{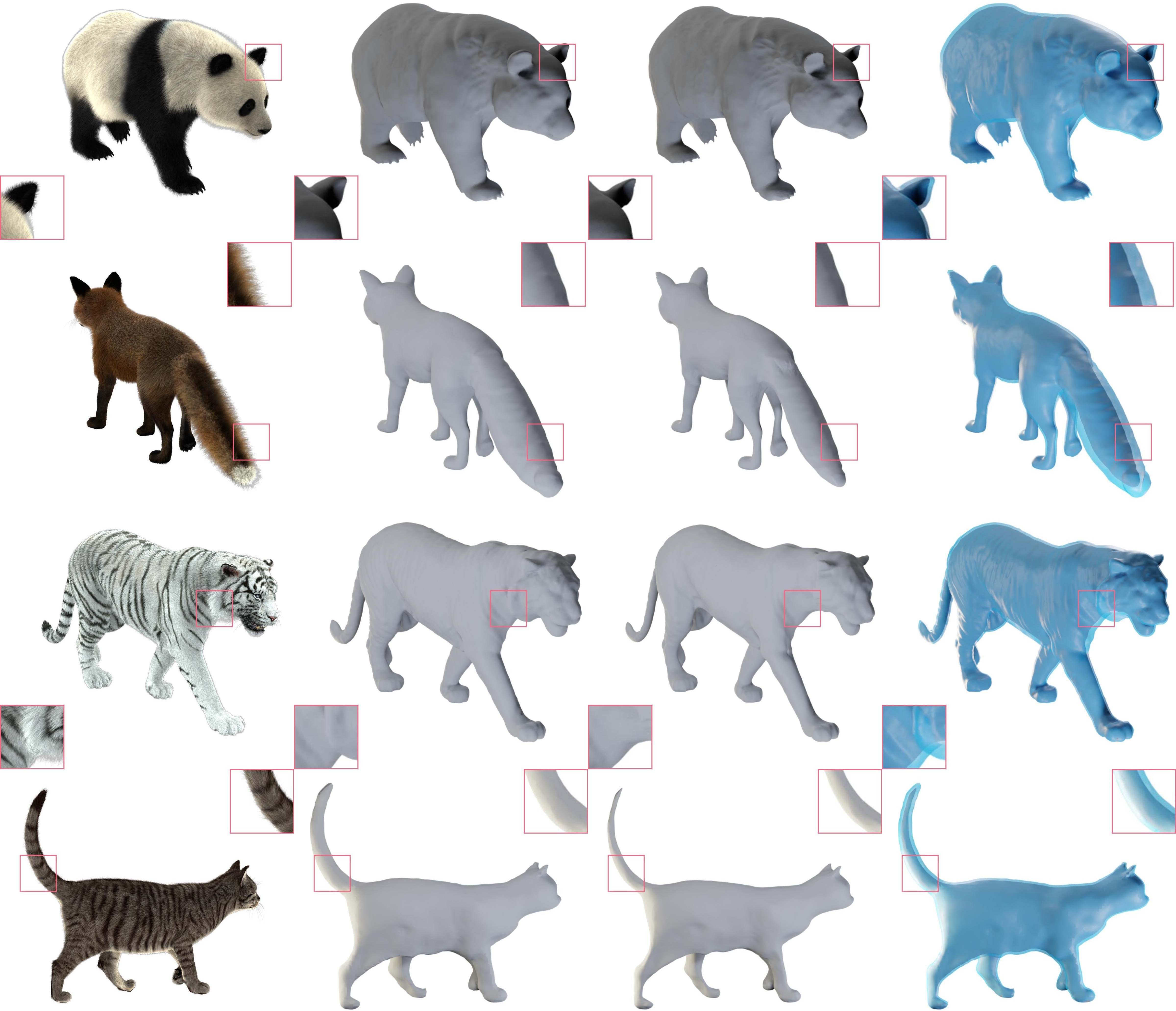} 
    \vspace{-0.25cm}
    \caption{
    From left to right: input image, initial shape $M_{NeuS}$, furless geometry $M_{bald}$, and their overlay. 
    }
    \label{fig:bald_geometry}
\end{figure}

%% file: author-kit/sec/experiments.tex
\section{Experiments}
\label{sec:experiments}
We train our method on the Artemis~\cite{artemis} dataset, which consists of 36 high-quality multi-view images captured for each animal.
Since no prior work addresses this specific task of animal reconstruction with fur modeling, we compare our approach against reconstruction approaches as well as a strand-based hair reconstruction method.
For quantitative evaluation, we use unsupervised metrics such as geometry consistency of the reconstructed fur and distance to the outer surface, and provide comparison results on an artist-created synthetic asset with ground-truth fur geometry.

\subsection{Qualitative evaluation}

\input{author-kit/figures/main_comparison}

We compare our method against popular 3D reconstruction approaches for animals, such as SMAL~\cite {Zuffi:CVPR:2017} and GenZoo~\cite{niewiadomski2024generativezoo}, general scene reconstructions, NeuS~\cite{Wang2021NeuSLN}, as well as a method designed for strand-based human hair reconstruction, Gaussian Haircut~\cite{zakharov2024gaussianhaircut}. 
As there are no prior works in the direction of fur modeling it serves as the closest baseline to ours.
We present qualitative results of our method with baselines in \Cref{fig:comparison}.
Note that SMAL, NeuS, and GenZoo are only able to produce coarse animal geometry, which includes the full outer surface.
While Gaussian Haircut (GH) generates realistic and high-quality reconstructions of strand-based hairstyles, it fails to reconstruct accurate fur geometry. In \Cref{fig:comparison}, we show results after the second stage, as the third stage significantly degrades performance. 
A key limitation is the lack of explicit control over strand length, which is only optimized implicitly.
In contrast, our method reconstructs realistic fur directly from images with high accuracy.

\subsection{Quantitative evaluation}

\input{author-kit/figures/main_ablation}
\input{author-kit/tables/ablation}

\input{author-kit/tables/ablation_real}

For quantitative comparison of our method with Gaussian Haircut~\cite{zakharov2024gaussianhaircut} and ablation of design choices of our model, we use an artist-created synthetic asset of a tiger with available ground-truth strand-based geometry for fur.
To do that, we realistically render the animal using Blender~\cite{Blender} from 90 views from a circular trajectory, and then use it as an input to the method.
Following ~\cite{Nam2019StrandAccurateMH, Rosu2022NeuralSL, sklyarova2023neural_haircut}, we measure precision, recall, and F-score between points with directions of reconstructed strands and the ground truth, see \Cref{tab:metrics_nh}.

We further evaluate our method on two scenes in terms of consistency of length, directions, and curvature in local (using ten nearest neighbours) and global spaces, and distance to the surface, see \Cref{tab:ablation_metrics_unsupervised}.
Specifically, we use unsupervised metrics that calculate:
(1) fur length statistics based on global length mean $\mu_{L}$ and standard deviation of strands $\sigma_L$ along with deviation in local region ($\sigma_{loc}(L)$), 
(2) fur curvature information based on local ($\mathrm{Var}_{\text{loc}}(\kappa)$ ) and global variance ($\mathrm{Var}_{\text{glob}}(\kappa)$ ) of curvature, and maximum bending angle $\kappa_{\max}$, 
(3) fur orientation based on local variance of directions for the whole strand ($\mathrm{Var}_{\text{loc}}(\text{dir})$) and considering only local variance of directions defined on first segment ($\mathrm{Var}_{\text{loc}}^{\text{first}}(\text{dir})$),
and (4) fur silhouette based on the two-way chamfer distance between the ground-truth geometry and the tips of the fur strands (CD).

\smallskip
\noindent\textit{Comparison with Gaussian Haircut.} As shown in \Cref{tab:metrics_nh,tab:ablation_metrics_unsupervised}, we compare results with Gaussian Haircut after the second and third stages (denoted as GH (2nd stage) and GH (3rd stage), respectively). We observe a notable degradation in quality compared to our method.

\smallskip
\noindent\textit{Part-specific fur length}. We ablate the importance of modeling different fur lengths for different parts, see \Cref{tab:metrics_nh,tab:ablation_metrics_unsupervised}.
In \Cref{fig:ablation_results}, we provide qualitative results of our method with fixed length (denoted as w/ fix. length), which exhibits artifacts especially when modeling long fur in the face region (see row 3). We also show results with implicitly optimized length, similar to Gaussian Haircut~\cite{zakharov2024gaussianhaircut} (w/ opt. length). However, unsupervised length optimization leads to unrealistic fur length and noticeable inconsistencies.

\smallskip
\noindent\textit{Accurate body geometry.} To demonstrate the importance of body geometry, we launch our reconstruction method with the body geometry replaced from $M_{bald}$ to $M_{NeuS}$, see  w/o defur in \Cref{fig:ablation_results}. While qualitatively it shows similar results, there is an increased level of unrealistic strands, as we could see in the decreased spatial consistency of directions and curvatures in \Cref{tab:ablation_metrics_unsupervised}. Finally, we measured the bidirectional Chamfer distance (in mm) between the ground-truth bald geometry $M_{bald}^{gt}$ and our reconstruction $M_{bald}$, as well as $M_{NeuS}$. The distances (mean / min / max) for $M_{bald}$: \textbf{6.83 / 0.45 / 69.72 mm} are significantly lower than for $M_{NeuS}$: \textbf{7.56 / 0.49 / 92.22 mm}.

\smallskip
\noindent\textit{Parametrization.} We explore different latent space parametrization by replacing our NeuralFur MLP with the Unet architecture from Neural Haircut~\cite{sklyarova2023neural_haircut}. From \Cref{fig:ablation_results} and \Cref{tab:metrics_nh,tab:ablation_metrics_unsupervised}, we can see significant degradation in consistency and realism of produced fur.

\smallskip
\noindent\textit{Prompting.} To model accurate responses, we requested fur lengths specifically in centimeters, asked for specific body parts, and provided images from several views. We evaluate the performance of length prediction on a synthetic asset, including ablations for predictions without images, with a single image, and with several images. The results are reported in Table~\ref{tab:ablation_gpt}.
For the structure of the prompt, please refer to the supplementary.

\input{author-kit/tables/gpt_ablation}

\smallskip
\noindent\textit{Losses.} 
We ablate the importance of chamfer, curvature, and orientation consistency losses in \Cref{tab:ablation_metrics_unsupervised}, see w/o $\mathcal{L}_{\mathrm{chm}}$, w/o $\mathcal{L}_{\mathrm{shape}}$, and w/o $\mathcal{L}^{gpt}_{\text{dir}}$ correspondingly. Across several animal models, omitting $\mathcal{L}_{\mathrm{chm}}$ results in degraded CD metric and reduced curvature consistency. Excluding $\mathcal{L}_{\mathrm{shape}}$ diminishes both local and global directional consistency. Finally, discarding $\mathcal{L}^{gpt}_{\text{dir}}$ yields greater inconsistencies in directional alignment as well as curvature.

\subsection{Discussion}
Our method relies solely on multi-view images and effectively reconstructs detailed fur grooms by leveraging complementary information retrieved from a VLM.
However, such multi-view data may not always be available for in-the-wild animals, and the results are dependent on the quality of the VLM outputs. 
Moreover, our framework is constrained by SMAL, limiting it to quadruped animals with a shared topology; extending the model to support a wider variety of animal types remains an exciting direction for future work.
Another promising direction is the explicit modeling of whiskers, which are essential features for many species, particularly, when close-up reconstructions of the face are required.
Finally, it is important to note that our approach is restricted to recovering static geometry; appearance attributes and animations must be incorporated afterwards within a Computer Graphics framework, see \Cref{fig:sumulations}.

\input{author-kit/figures/figure_sim}

%% file: author-kit/figures/main_comparison.tex
\begin{figure*}[t]
    \centering
    \includegraphics[width=\linewidth, trim=0 0 0 23cm, clip]{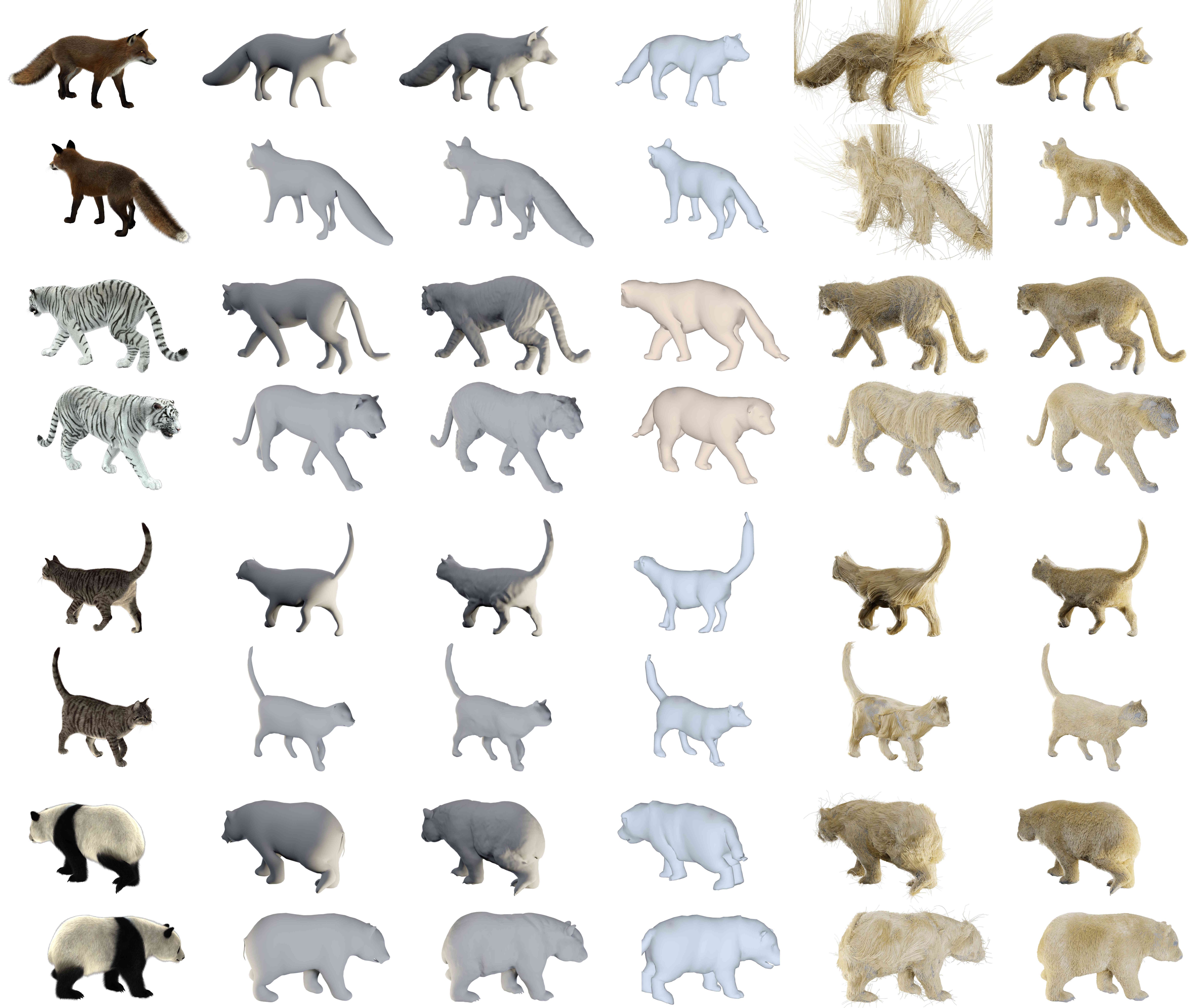}
    \begin{tabular*}{0.9\linewidth}{@{\extracolsep{\fill}}l l l l l l}
Image & SMAL~\cite{Zuffi:CVPR:2017} & NeuS~\cite{Wang2021NeuSLN} & GenZoo~\cite{niewiadomski2024generativezoo} & GH~\cite{zakharov2024gaussianhaircut} & Ours \\
\end{tabular*}
    
    \caption{Qualitative results of our reconstruction method compared with existing baselines. Surface reconstruction baselines produce very coarse geometry. Applying the existing state-of-the-art hair reconstruction method, Gaussian Haircut~\cite{zakharov2024gaussianhaircut} leads to inconsistent strand lengths and noticeable artifacts. Our method produces accurate strand-based geometry. A digital zoom-in is recommended. 
    }
    \label{fig:comparison}
\end{figure*}

%% file: author-kit/figures/main_ablation.tex
\begin{figure*}[t]
    \centering
    \includegraphics[width=\linewidth, trim=0 0 0 0cm, clip]{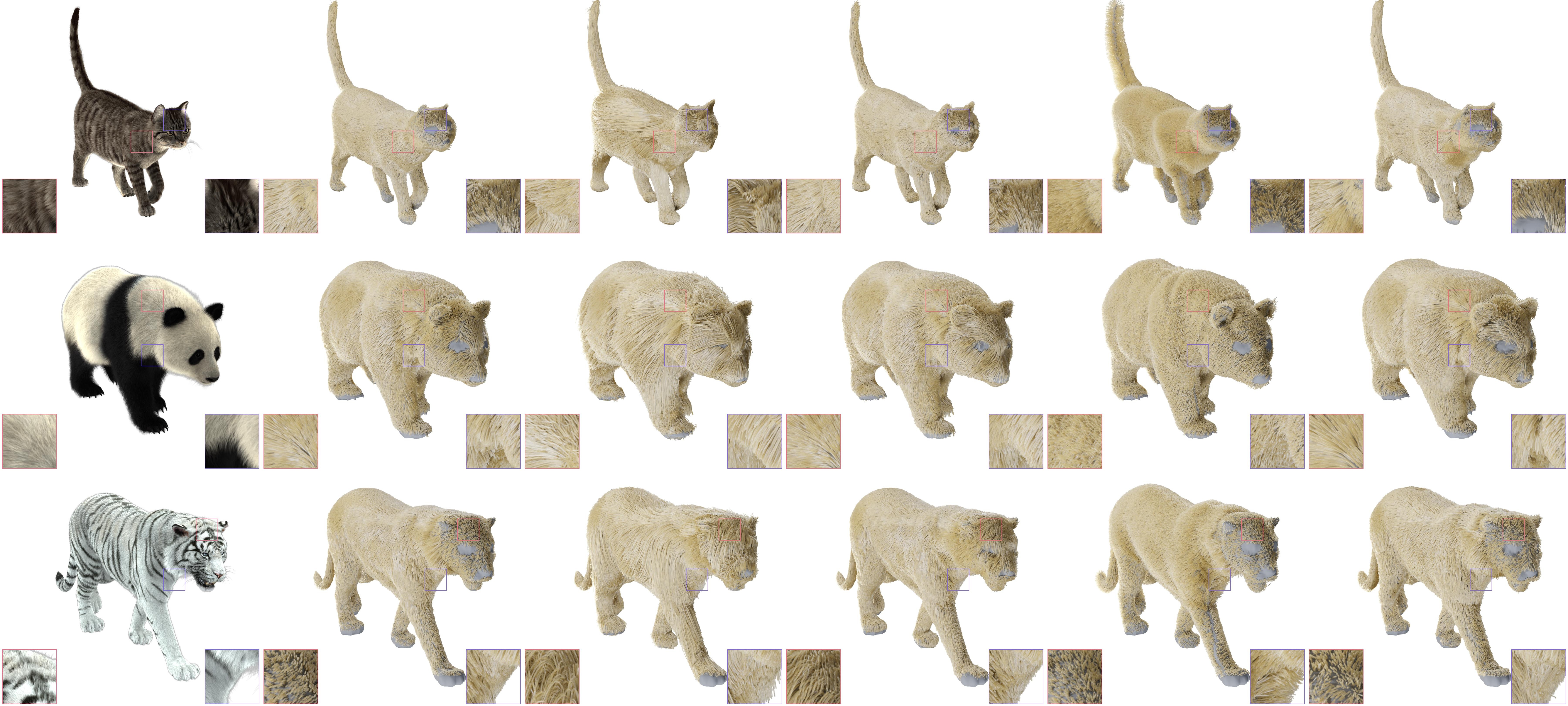}
    \vspace{0.25cm}
    \makebox[0.162\linewidth]{Image}%
    \makebox[0.162\linewidth]{Ours}%
    \makebox[0.162\linewidth]{w/ opt. length}%
    \makebox[0.162\linewidth]{w/ fix. length}
    \makebox[0.162\linewidth]{ w/ U-Net}%
    \makebox[0.162\linewidth]{w/o defur }%
    \vspace{-0.25cm}
    \caption{ \textbf{Ablation study.} Qualitative evaluation of our design choices regarding length, fur parametrization, and the importance of the defurring approach for accurate geometry modeling.}
    \label{fig:ablation_results}
\end{figure*}

%% file: author-kit/tables/ablation.tex
\begin{table}[t!]
\centering
    \centering
    \resizebox{0.99\linewidth}{!}{
    \begin{tabular}{l rrr | rrr | rrr}
        \setlength{\tabcolsep}{0pt}
        & \multicolumn{9}{c}{\textbf{Thresholds: cm} $/$ \textbf{degrees}} \\
        \textbf{Method} & $2 / 20$ & $3 / 30$ & $4 / 40$ & $2 / 20$ & $3 / 30$ & $4 / 40$ & $2 / 20$ & $3 / 30$ & $4 / 40$ \\
        \cline{2-10}
        & \multicolumn{3}{c}{\textbf{Precision}} & \multicolumn{3}{c}{\textbf{Recall}} & \multicolumn{3}{c}{\textbf{F-score}} \\
        \hline
\hline
Ours           &                      26.22 &                      39.32 &                      48.05 &                      20.58 &                      34.08 &                      45.69 &                      23.06 &                      36.51 &                      46.84 \\
GH (2nd stage) &                      16.24 &                      25.51 &                      32.34 &                      23.51 &                      36.04 &                      45.87 &                      19.21 &                      29.87 &                      37.93 \\
GH (3rd stage) &  5.78 &  10.44 &  15.43 &                      21.39 &  32.82 &  42.36 &  9.10 &  15.84 &  22.62 \\
w/ opt.length  &  3.05 &  6.26 &  9.82 &  16.63 &  28.27 &  38.45 &  5.16 &  10.25 &  15.65 \\
w/ fix. length &                      17.18 &                      26.65 &                      33.73 &                      22.24 &                      35.08 &                      45.44 &                      19.38 &                      30.29 &                      38.72 \\
w/ Unet        &  5.77 &  12.16 &  20.06 &  7.05 &  15.39 &  25.80 &  6.35 &  13.58 &  22.57 \\

    \end{tabular}
    }
    \captionof{table}{Quantitative evaluation on a synthetic tiger model where ground truth information of the fur is given as reference.}%
    \label{tab:metrics_nh}
\end{table}

%% file: author-kit/tables/ablation_real.tex
\definecolor{tabfirst}{rgb}{1, 0.7, 0.7} %
\definecolor{tabsecond}{rgb}{1, 0.85, 0.7} %
\definecolor{tabthird}{rgb}{1, 1, 0.7} %

\begin{table}[tbh]
\centering
\setlength{\tabcolsep}{4pt}  %
\renewcommand{\arraystretch}{1.2} %
\resizebox{\linewidth}{!}{
\begin{tabular}{l|cc|ccc|cc|c}
 & \multicolumn{8}{c}{\textit{``Panda''}} \\
\hline
Method & 
\multicolumn{2}{c|}{Fur length} & 
\multicolumn{3}{c|}{Fur curvature} &
\multicolumn{2}{c|}{Fur orientation} &
Fur silhouette \\
& $\mu_{L}$ $\pm$ $\sigma_L$ (cm)
& $\sigma_{\text{loc}}(L)$ $\downarrow$ 
& $\mathrm{Var}_{\text{glob}}(\kappa) \downarrow$
& $\mathrm{Var}_{\text{loc}}(\kappa) \downarrow$
& $\kappa_{\max} \downarrow$
& $\mathrm{Var}_{\text{loc}}(\text{dir}) \downarrow$
& $\mathrm{Var}_{\text{loc}}^{\text{first}}(\text{dir}) \downarrow$
& CD $\downarrow$\\

\hline
Ours                  & \cellcolor{tabsecond}5.197 $\pm$ 1.38 & \cellcolor{tabsecond}0.283 &  \cellcolor{tabfirst}0.0005 &  \cellcolor{tabfirst}0.000069 & \cellcolor{tabsecond}1.80 &  \cellcolor{tabfirst}0.031 & \cellcolor{tabsecond}0.041 &  \cellcolor{tabthird}0.000261 \\
GH (2nd stage)      &                      7.322 $\pm$ 8.06 &                      3.615 &                      0.0018 &                      0.000789 &                      3.10 &                      0.317 &                      0.302 &                      0.001105 \\
GH (3rd stage) &                      11.238 $\pm$ 7.53 &                      4.362 &                      0.2318 &                      0.352237 &                      3.13 &                      0.730 &                      0.765 &                      0.001600 \\
w/ opt. length            &                      17.169 $\pm$ 10.59 &                      1.469 &                      0.0048 &                      0.000681 &                      3.11 &  \cellcolor{tabthird}0.035 &                      0.054 &                      0.000637 \\
w/ fix. length            &  \cellcolor{tabthird}6.000 $\pm$ 0.00 &  \cellcolor{tabfirst}0.000 &                      0.0014 &  \cellcolor{tabthird}0.000192 &                      3.05 &  \cellcolor{tabthird}0.035 &                      0.053 & \cellcolor{tabsecond}0.000257 \\
w/ Unet           &  \cellcolor{tabfirst}5.196 $\pm$ 1.38 &  \cellcolor{tabthird}0.285 & \cellcolor{tabsecond}0.0007 &                      0.001282 &  \cellcolor{tabfirst}1.35 &                      0.157 &                      0.200 &  \cellcolor{tabfirst}0.000254 \\
w/o defur             & \cellcolor{tabsecond}5.197 $\pm$ 1.35 &                      0.291 &                      0.0019 &                      0.000329 &                      2.90 &                      0.044 &                      0.060 &                      0.000802 \\
w/o $\mathcal{L}_{\mathrm{chm}}$             & \cellcolor{tabsecond}5.197 $\pm$ 1.38 & \cellcolor{tabsecond}0.283 &  \cellcolor{tabthird}0.0010 & \cellcolor{tabsecond}0.000149 &  \cellcolor{tabthird}2.65 & \cellcolor{tabsecond}0.033 &  \cellcolor{tabfirst}0.033 &                      0.000330 \\
w/o $\mathcal{L}^{gpt}_{\text{dir}}$               & \cellcolor{tabsecond}5.197 $\pm$ 1.38 & \cellcolor{tabsecond}0.283 &                      0.0012 &                      0.000224 &                      2.71 &                      0.041 &  \cellcolor{tabthird}0.050 &  \cellcolor{tabthird}0.000261 \\
w/o $\mathcal{L}_{\mathrm{shape}}$              &  \cellcolor{tabfirst}5.196 $\pm$ 1.38 & \cellcolor{tabsecond}0.283 &                      0.0016 &                      0.000284 &                      2.91 & \cellcolor{tabsecond}0.033 &                      0.053 &                      0.000265 \\
\hline
\hline
& \multicolumn{8}{c}{\textit{``White Tiger''}} \\
\hline
Ours                  &  \cellcolor{tabfirst}3.555 $\pm$ 1.46 & \cellcolor{tabsecond}0.355 & \cellcolor{tabsecond}0.0016 &  \cellcolor{tabfirst}0.000210 &                      3.05 &  \cellcolor{tabfirst}0.030 &  \cellcolor{tabfirst}0.035 &  \cellcolor{tabthird}0.000207 \\
GH (2nd stage)      &                      6.511 $\pm$ 17.86 &                      8.190 &                      0.0144 &                      0.006519 &                      3.12 &                      0.526 &                      0.469 &                      0.000312 \\
GH (3rd stage)  &                      9.929 $\pm$ 17.24 &                      8.087 &                      0.4049 &                      0.636722 &                      3.14 &                      0.885 &                      0.970 &                      0.000553 \\
w/ opt. length            &                      19.992 $\pm$ 15.20 &                      1.756 &                      0.0047 &                      0.000637 &                      3.09 &  \cellcolor{tabthird}0.036 &                      0.050 &                      0.000453 \\
w/ fix. length            &                      5.000 $\pm$ 0.00 &  \cellcolor{tabfirst}0.000 &                      0.0023 &  \cellcolor{tabthird}0.000352 &                      3.06 &                      0.037 &                      0.050 & \cellcolor{tabsecond}0.000202 \\
w/ Unet           & \cellcolor{tabsecond}3.556 $\pm$ 1.46 & \cellcolor{tabsecond}0.355 &  \cellcolor{tabfirst}0.0008 &                      0.001337 &  \cellcolor{tabfirst}1.39 &                      0.124 &                      0.081 &                      0.000379 \\
w/o defur             &  \cellcolor{tabthird}3.570 $\pm$ 1.48 &  \cellcolor{tabthird}0.357 &                      0.0036 &                      0.000728 &  \cellcolor{tabthird}2.96 &                      0.051 &                      0.082 &                      0.000420 \\
w/o $\mathcal{L}_{\mathrm{chm}}$             &  \cellcolor{tabfirst}3.555 $\pm$ 1.46 & \cellcolor{tabsecond}0.355 &  \cellcolor{tabthird}0.0021 & \cellcolor{tabsecond}0.000306 & \cellcolor{tabsecond}2.66 & \cellcolor{tabsecond}0.031 &  \cellcolor{tabthird}0.040 &                      0.000258 \\
w/o $\mathcal{L}^{gpt}_{\text{dir}}$               &  \cellcolor{tabfirst}3.555 $\pm$ 1.46 & \cellcolor{tabsecond}0.355 &                      0.0041 &                      0.000862 &                      3.04 &                      0.049 & \cellcolor{tabsecond}0.037 &  \cellcolor{tabfirst}0.000190 \\
w/o $\mathcal{L}_{\mathrm{shape}}$              &  \cellcolor{tabfirst}3.555 $\pm$ 1.46 & \cellcolor{tabsecond}0.355 &                      0.0034 &                      0.000537 &                      3.08 &  \cellcolor{tabfirst}0.030 &                      0.056 & \cellcolor{tabsecond}0.000202 \\

\hline
\end{tabular}
}
\caption{Unsupervised geometry consistency metrics for length, direction, and curvature, evaluated for both local and global cases, as well as for coverage of the outer surface. 
}
\label{tab:ablation_metrics_unsupervised}
\end{table}

%% file: author-kit/tables/gpt_ablation.tex
\begin{table}[t]
    \centering
    \resizebox{0.92\linewidth}{!}{
    \begin{tabular}{l|c|ccc}
&Ground Truth (cm)& \multicolumn{3}{c}{GPT responses (cm)} \\
\cline{3-5}
 & (mean / min / max) & No image & 1 image & 2 images\\
    \hline
Face &0.61 / 0.00008 / 2.57 &  1 - 2 &                      1 - 1.5&  0.5 - 1\\ %
Body &2.30 / 0.0011 / 8.60 &  2 - 5 &                      2 - 3 & 1 - 2 \\
Legs &0.65 / 0.15 / 1.26 &  2 - 4 &                      1.5 - 2.5 &  0.5-1.5\\ %
Mane &1.89 / 0.00 / 9.27 & 5 - 8 &                      6 - 7 &  3 - 5 \\
\hline
    \end{tabular}
    }
    \caption{Comparison of results on length estimation for different parts obtained from ChatGPT on a synthetic asset.}
    \label{tab:ablation_gpt}
\end{table}

%% file: author-kit/figures/figure_sim.tex
\begin{figure}[t]
    \centering
    \vspace{-0.125cm}
    \includegraphics[width=0.85\linewidth]{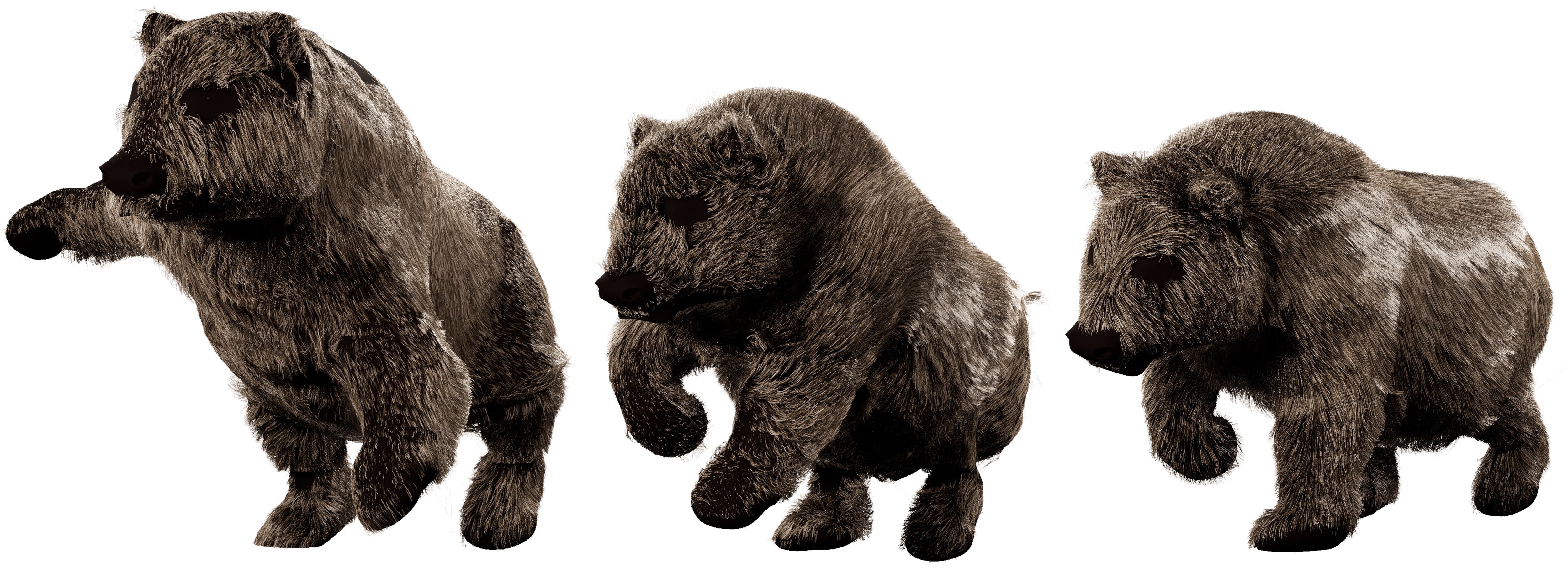} 
    \vspace{-0.1cm}
    \caption{
    Simulation of a reconstructed panda with fur in Unreal Engine~\cite{unrealengine}. Note that the fur color is set by an artist, and the animation is applied via the SMAL model.
    }
    \label{fig:sumulations}
\end{figure}

%% file: author-kit/sec/conclusion.tex
\section{Conclusion}
\label{sec:conclusion}
In this work, we introduced the first method capable of reconstructing strand-based fur geometry from multi-view images.
Our method integrates furless geometry extraction with a subsequent fur optimization stage, leveraging prior knowledge from VLMs to estimate fur length across body regions and hair growth direction, resulting in notable improvements in reconstruction quality.
We argue that leveraging external knowledge sources such as VLMs is particularly valuable for 3D reconstruction tasks where data is scarce and explicit priors cannot be easily learned.

%% file: author-kit/sec/acknowledgement.tex
\section*{Acknowledgements}
Vanessa Sklyarova is supported by the Max Planck ETH Center for Learning Systems. Berna Kabadayi is supported by the International Max Planck Research School for Intelligent Systems (IMPRS-IS). Justus Thies is supported by the ERC Starting Grant 101162081 ``LeMo'' and the DFG Excellence Strategy— EXC-3057. 
The authors would like to thank Peter Kulits and Silvia Zuffi for their discussions on the project, Tomasz Niewiadomski for providing results on GenZoo, and Benjamin Pellkofer for IT support.

\paragraph{Disclosure.} 
While MJB is a co-founder and Chief Scientist at Meshcapade, his research in this project was performed solely at, and funded solely by, the Max Planck Society.

%% file: author-kit/supp_sec/experiments.tex
\section{Preliminaries}
\input{author-kit/sec/background}

\section{Training details}

\subsection{Strands parametrization}

\textit{Basis calculation.} We map each strand into a local tangent-bitangent-normal (TBN) basis using the vertices from the closest face to its root location on the $M_{bald}$ mesh. %
Each strand consists of $L=100$ points. The normal is computed on the mesh and is consistent with nearby strands. In contrast, the definition of a tangent vector is problematic. Different to prior works on hair reconstruction~\cite{Rosu2022NeuralSL, sklyarova2023neural_haircut, zakharov2024gaussianhaircut}, we align tangents with the computed directional face field with further sign consistency resolving using parallel transport~\cite{spivak1979comprehensive} across shared edges. The bitangent vector is then defined as a cross-product between the normal and the tangent.
For directional field estimation, we use the library \url{https://github.com/avaxman/Directional}.

\subsection{Optimization details}

We launch the NeuS~\cite{Wang2021NeuSLN} reconstruction method for 300,000 iterations, which takes around 6 hours on a single A100.
For SMAL fitting, we use the Adam optimizer~\cite{Adam} with different learning rates depending on the optimization stage. Following WLDO~\cite{biggs2020wldo}, we minimize the following energy function:
\begin{equation}
\begin{aligned}
L_{\text{total}} &= L_{\text{chamfer}} 
+ \mathcal{L}_{\mathrm{laplacian}} L_{\text{laplacian}} \\
&\quad + \mathcal{L}_{\mathrm{edge}} L_{\text{edge}} 
+ \mathcal{L}_{\mathrm{normal}} L_{\text{normal}},
\end{aligned}
\end{equation}
where $\mathcal{L}_{\mathrm{laplacian}}=0.01, ~\mathcal{L}_{\mathrm{edge}}=0.8, ~\mathcal{L}_{\mathrm{normal}}=0.02.$
For the final optimization of our model, we use the following weights for losses: 
$\mathcal{L}_{\mathrm{sil}}=0.1,~\mathcal{L}_{\mathrm{dir}}=1000,~\mathcal{L}^{gpt}_{\mathrm{dir}}=1,~ \mathcal{L}_{\mathrm{chm}}=20,~ \mathcal{L}_{\mathrm{penetr}}=1,~ \mathcal{L}_{\mathrm{shape}}=0.01$ and strand width 0.0025. We use $\gamma=10$ for positional encodings.

\subsection{De-furring process}

For the de-furring process, we convert the mesh to an SDF. We first estimate the effective strand thickness for each region using ChatGPT, then smooth these values via neighborhood averaging. The mesh is converted to an SDF, shrunk according to the predicted thickness values, and converted back using marching cubes with resolution 256. Finally, we resample the resulting mesh to 160,000 faces and apply basic mesh repair operations to remove noise and artifacts.

\subsection{Prompting}

To obtain annotations of an animal, we sent two images from frontal and side views and then provided a set of prompts.

\smallskip
\noindent\textit{Length annotations.}  We annotate the fur length using the following prompt:
\textcolor{gray}{``Here, you could see images of an animal. Could you estimate the accurate fur length in cm for each part of the animal: ``leg\_front'', ``leg\_rear'', ``paws'',  ``front\_paws'',   ``belly'',   ``neck'',    ``face'',   ``ears'',      ``under\_tail'',  ``tail'', ``body'', ``paw\_pads'', ``inner\_earcanal'', ``eyes'', ``nosetip''? Also, does this animal have some fur near the neck that grows significantly beyond the underlying body? If so, could you add the estimates for length and fur thickness as well for ``mane''  and include it at the end of the previous part names; otherwise, do not include it.  Is the ear canal visible in the image? If the inner ear canal is not visible, use the same value for length as for the outer ear. Please provide results in dict format, where the key is part name and the value is your estimate.''}

\smallskip
\noindent\textit{Effective fur thickness.}To determine the effective fur thickness for the subsequent de-furring step, we employ the following prompt: 
\textcolor{gray}{``I want to create a furless animal. To do that, I have a 3D model of an animal with fur, from which I want to subtract the region covered by fur for each part. Could you provide the number of effective fur thickness for each part that I need to subtract from the full geometry? Please provide results in dict format where the key is part name and the value is your estimate.''} 
Finally, we verify that the estimates are correct and correspond to the image using the following prompt:
\textcolor{gray}{``Are you sure of the initial length estimations and fur thickness to abstract? Could you double-check with the image? Please provide results in a several-dict format where the key is part name and the value is your estimate.''}

\smallskip
\noindent\textit{Hair growing direction}. To determine the hair growth direction, we provide the coordinate space, specify its relation to the animal, and then ask:
\textcolor{gray}{``Could you also estimate what the fur growing direction looks like in 3d space? I want to know which parts are oriented along the gravity vector and which are against. Also, if fur for regions grows from left to right or from right to left. I have the following coordinate system: x from the right side of animal towards the left, y - opposite to gravity direction, and z from back to the front of the animal. The second image is aligned with the 3D coordinate system. I want for each part: ``leg\_front'', ``leg\_rear'', ``paws'',  ``front\_paws'',   ``belly'',   ``neck'',    ``face'',   ``ears'',      ``under\_tail'',  ``tail'', ``body'', ``paw\_pads'', ``inner\_earcanal'', ``eyes'', ``nosetip'', ``mane'' (if appeared) obtains a vector that defines the approximate growing direction in the coordinate system. Please double-check the proposed directions several times.''}

\smallskip
\noindent\textit{Eyeballs annotations.} To annotate the distance between the centers of the eyes, we employ the following prompt:
\textcolor{gray}{``Could you measure the distance between eyeballs in cm? You could also use the real-world  understanding of this animal class.''}

\medskip
The final length annotations for each animal part obtained from ChatGPT are shown in centimeters in Table~\ref{tab:mapping_lengths}. Figure~\ref{fig:length-annotation} illustrates the distribution of fur lengths across all animals, while the distribution of effective fur thickness is shown in Figure~\ref{fig:effective-length-annotation}.

\input{author-kit/tables_suppmat/length_info}

\input{author-kit/figures_supmat/length_anno}

\input{author-kit/figures_supmat/effective_length}

\subsection{Data preprocessing}

We use multi-view images of animals with ground-truth silhouettes, along with camera parameters provided in the Artemis~\cite{artemis} dataset. After obtaining a full mesh reconstruction from NeuS~\cite{Wang2021NeuSLN}, denoted as $M_{\text{NeuS}}$, we fit SMAL to it to transfer body part labels. Next, we use ChatGPT to obtain effective length and thickness annotations for each part. Based on this information, we reconstruct the de-furred geometry $M_{\text{bald}}$. We then re-fit SMAL to $M_{\text{bald}}$ to obtain vertex-level length and direction annotations, which are subsequently used during the optimization process. Finally, we compute the SDF field for $M_{\text{bald}}$, which is required for the penetration loss.

\subsection{Baselines}

The GenZoo~\cite{niewiadomski2024generativezoo} results were provided by the corresponding authors, while NeuS~\cite{Wang2021NeuSLN}, AniMer~\cite{lyu2025animer}, and GaussianHaircut~\cite{zakharov2024gaussianhaircut} were run using their respective public repositories.

\subsection{Datasets}

\input{author-kit/figures_supmat/synth_data}

Our model is evaluated on five scenes from the Artemis~\cite{artemis} dataset. For quantitative evaluation we use synthetic asset with ground-truth geometry from \url{https://superhivemarket.com/products/complete-realistic-tiger---texture--fur--armature?search_id=41647510}. Figure~\ref{fig:synth} illustrates several views generated by rendering the scene in Blender~\cite{Blender}.

\subsection{Metrics}

For quantitative evaluation on a synthetic asset, we compute metrics using 100,000 strands sampled from both the ground-truth and generated assets. While the reconstructed furred animal appears visually close to the ground truth, we observe relatively high metric values due to (1) inaccuracies in the defurring procedure and (2) inaccuracies in the strand length prediction process from VLM. Including additional synthetic assets for benchmarking is a potential direction for future work.

To assess the consistency of the generated strands, we compute self-supervised metrics for local and global consistency in terms of length, direction, and curvature. 
Let M be the number of points per strand. To compute the length of each strand, we use the following equation:
\begin{equation}
L_i = \sum_{l=0}^{M-2} \| \mathbf{p}_{i}^{l+1} - \mathbf{p}_{i}^{l} \|_2 .
\end{equation}
We calculate the strand’s average tangent direction by:
\begin{align}
\tilde{\mathbf{t}}_i &= \frac{1}{M-1} \sum_{l=0}^{M-2} \mathbf{b}_i^l  \\
\mathbf{t}_i &= \frac{\tilde{\mathbf{t}}_i}{\|\tilde{\mathbf{t}}_i\|_2} .
\end{align}

\medskip
\noindent
Based on these measurements, we define the following metrics:

\smallskip
\noindent\textit{Local length variance:}
The local length variance is defined as follows:
\begin{equation}
\sigma_{loc}(L) = 
\sqrt{ \frac{1}{N} \sum_{i=1}^N \frac{1}{k} \sum_{j \in \text{knn}(i)} (L_i - L_j)^2 } .
\end{equation}

\smallskip
\noindent\textit{Global curvature variance:}
The global curvature variance is defined analogously to the optimization loss in the main paper: $\mathrm{Var}_{\text{glob}}(\kappa) \equiv \mathcal{L}_{\mathrm{shape}}$.

\smallskip
\noindent\textit{Local curvature variance:}
The local curvature variance is formulated in a similar manner, but restricted to the $k$ nearest neighbors:
\begin{equation}
\mathrm{Var}_{\text{loc}}(\kappa) = 
\frac{1}{N} \sum_{i=1}^N \frac{1}{k} \sum_{j \in \text{knn}(i)} 
\frac{1}{M-2} \sum_{l=1}^{M-2} (\theta_{i}^{l} - \theta_{j}^{l})^2 .
\end{equation}

\smallskip
\noindent\textit{Local direction consistency:}
Similarly, we define local direction consistency. We first evaluate the consistency score based on the initial segment directions:
\begin{equation}
\mathrm{Var}_{\text{loc}}^{\text{first}}(\text{dir}) = \frac{1}{N} \sum_{i=1}^N \frac{1}{k} \sum_{j \in \text{knn}(i)} \| \mathbf{b}^0_i - \mathbf{b}^0_j \|_2^2 .
\end{equation}

\smallskip
\noindent\textit{Local direction variance:}
The local direction variance is calculated as follows:
\begin{equation}
\mathrm{Var}_{\text{loc}}(\text{dir}) = \frac{1}{N} \sum_{i=1}^N \frac{1}{k} \sum_{j \in \text{knn}(i)} \| \mathbf{t}_i - \mathbf{t}_j \|_2^2 .
\end{equation}

\smallskip
\noindent\textit{Maximum bending angle:}
Lastly, the maximum bending angle is given by the following expression:
\begin{equation}
\kappa_{\max} = \max\limits_{i,l} \theta_i^l .
\end{equation}

\section{Applications}

\noindent\textbf{Rendering.} For rendering in Unreal Engine~\cite{unrealengine}, we use 500,000 strands. Figure~\ref{fig:unreal-rendering} shows the results of importing reconstructed furless animals and fur into Unreal Engine and rendering them with the specified colors.

~

\noindent\textbf{Simulations.} We simulate the reconstructed fur using the fitted SMAL~\cite{Zuffi:CVPR:2017} model along with its kinematic tree. Additional results are provided in the Supplementary Video.

~

\noindent\textbf{Editing.} Our 3D part-based annotations allow region-specific editing, providing fine-grained control over fur on different body areas. For example, in Figure~\ref{fig:editing_suppmat}, we show edits where the tail fur length is changed from 6 cm to 12 cm, leg fur from 4 cm to 8 cm, and face fur from 3 cm to 5 cm, while fur on other regions remains unchanged. More results are available in the Supplementary Video.

\input{author-kit/figures_supmat/editing_fur}

\section{Additional experiments}

\subsection{Reconstructed results}

In Figure~\ref{fig:comparison_suppmat_animals}, we present additional results comparing our method with coarse reconstruction baselines, including SMAL~\cite{Zuffi:CVPR:2017}, GenZoo~\cite{niewiadomski2024generativezoo}, AniMer~\cite{lyu2025animer} and NeuS~\cite{Wang2021NeuSLN}. We also compare against the strand-based hair reconstruction method, Gaussian Haircut~\cite{zakharov2024gaussianhaircut}, as shown in Figure~\ref{fig:comparison_suppmat_GH}. For Gaussian Haircut, comparisons are provided after the second and third stages, denoted as GH (2nd stage) and GH (3rd stage), respectively.

\subsection{Extended ablation}

\noindent\textbf{Quantitative and qualitative results.} In Figure~\ref{fig:ablation_results_supp}, we present an additional ablation study examining strand length, the strand encoder, and the impact of the de-furring strategy. Table~\ref{tab:ablation_metrics_unsupervised_suppmat} reports self-supervised metrics on an additional scene.

\input{author-kit/tables_suppmat/ablation_real_suppmat}

\noindent\textbf{Number of views.} In Figure~\ref{fig:ablation_results_supp_num_views}, we present an ablation study on the number of views used for supervision during the second stage of strand-based fur optimization. Results are shown for 1, 2, 4, 8, and 16 input views, compared against the full 36-view setting. Note that all results are rendered from novel viewpoints.

\section{Limitations}
\input{author-kit/figures_supmat/limitation_dog}

\input{author-kit/figures_supmat/ablations_num_views}

Our goal is to automatically annotate parts of various animals by fitting a parametric SMAL~\cite{Zuffi:CVPR:2017} model. Nevertheless, errors in the fitted SMAL model may result in incorrect annotations, such as the dog ear being labeled as ``face'' in Figure~\ref{fig:dog-limitation}. Still, our method provides explicit control over fur growth: it does not grow any fur in regions that naturally lack fur, such as paw pads or nose tips (see Figure~\ref{fig:comparison_suppmat_animals}), showing that it might potentially generalize to hairless or partially hairless species. Nonetheless, its applicability is constrained by the SMAL model itself, which represents quadrupeds and cannot accurately represent animals with very different body structures (e.g., ducks or crocodiles). One possible solution to reconstruct animal fur without relying on SMAL is to bypass SMAL entirely during annotation and instead annotate parts directly on the reconstructed NeuS geometry. Although this limits the automatic transfer of fur length to corresponding body parts, such an approach would eliminate the dependency on a parametric model template and potentially might extend fur reconstruction to a broader range of animal species.   

\input{author-kit/figures_supmat/ablations_supp}

\input{author-kit/figures_supmat/rendering}

\input{author-kit/figures_supmat/comparison_baselines_supp}

\input{author-kit/figures_supmat/comparison_baselines_supp_GH}

%% file: author-kit/sec/background.tex
\paragraph{Parametric Animal Model.}

SMAL~\cite{Zuffi:CVPR:2017} is a parametric model for quadruped animals. Formally, it is a function \( M(\boldsymbol{\beta}, \boldsymbol{\theta}, \boldsymbol{\gamma}) \) that outputs a posed 3D animal shape, given shape parameters \( \boldsymbol{\beta} \), pose parameters \( \boldsymbol{\theta} \), and global translation \( \boldsymbol{\gamma} \).
Template vertices \(\mathbf{v}_t \in \mathbb{R}^{3889 \times 3}\) are transformed using specified shape parameters \(\boldsymbol{\beta}\) and learned PCA shape space \(\mathbf{B}\), then articulated using \(\boldsymbol{\theta}\) through LBS with joint regressor \(\mathbf{J}_r\) and LBS weights \(\boldsymbol{W}\):
\[
\mathbf{v}_d = \text{LBS}(\mathbf{v}_t + \boldsymbol{\beta} \cdot \mathbf{B},\ \boldsymbol{\theta};\ \boldsymbol{W},\ \mathbf{J}_r) + \boldsymbol{\gamma} .
\]
\(\boldsymbol{\theta}\) represents the relative rotations of joints in the kinematic tree. Translation \(\boldsymbol{\gamma}\) is applied to the root joint. 
\paragraph{Gaussian Splatting.}
To use 3D Gaussian Splatting for soft-rasterization of fur, we force Gaussians to lie on line segments.
We define the mean and covariance matrix of each individual Gaussian as:
\begin{equation}
    \mu_{i}^{l} = \frac{1}{2} \big( \p^{l}_i + \p^{l+1}_i \big),\quad C_{i}^{l} = E_{i}^{l} D_{i}^{l} \big ( E_{i}^{l} D_{i}^{l} \big)^T.
\end{equation}
Here, $E_{i}^{l} = \{ b_{i}^{l}, t_{i}^{l}, n_{i}^{l} \}$ is a TBN basis associated with the strand curve, $\mathbf{b}^l_i = \mathbf{d}^l_i \big/ \|\mathbf{d}^l_i\|_2$, where $\mathbf{d}^l_i =  \p^{l+1}_i-\p^l_i$ denotes the segment vector and $\mathbf{b}^l_i$ its normalized direction vector. $D_{i}^{l}$ is defined as $D_{i}^{l} = \text{diag} ( \mathbf{f}^l_i, \epsilon, \epsilon )$, where $\mathbf{f}^l_i$ is set to be proportional to the length of $\mathbf{d}^l_i$ and $\epsilon$ denotes a small value.
Such parametrization allows effective propagation of photometric information into fur geometry.

%% file: author-kit/tables_suppmat/length_info.tex
\begin{table}[h!]
\centering
\centering
\setlength{\tabcolsep}{4pt}  %
\renewcommand{\arraystretch}{1.2} %
\resizebox{\linewidth}{!}{
\begin{tabular}{lccccc}
\hline
Body Part & ``Cat'' & ``Beagle dog'' & ``Fox'' & ``Panda'' & ``White Tiger'' \\
\hline
``leg\_front'' & 0.8 & 0.7 & 1.35 & 4 & 2.1 \\
``leg\_rear'' & 0.8 & 0.7 & 1.75 & 4 & 2.3 \\
``paw\_pads'' & 0.1 & 0.0 & 0 & 0 & 0 \\
``paws'' & 0.45 & 0.4 & 0.6 & 3 & 1.1 \\
``front\_paws'' & 0.45 & 0.4 & 0.6 & 3 & 1.1 \\
``belly'' & 1.25 & 1.2 & 3.5 & 7.5 & 6 \\
``neck'' & 1.2 & 1.5 & 4.5 & 7 & 5 \\
``face'' & 0.5 & 0.4 & 1.1 & 3 & 1.35 \\
``ears'' & 0.3 & 1.8 & 0.6 & 4 & 1.5 \\
``inner\_earcanal'' & 0.15 & 1.8 & 0.25 & 0.5 & 0.4 \\
``under\_tail'' & 1 & 1.2 & 5 & 5 & 3 \\
``eyes'' & 0 & 0 & 0 & 0 & 0 \\
``tail'' & 1.35 & 1.8 & 7.8 & 6 & 3.25 \\
``nosetip'' & 0 & 0.0 & 0 & 0 & 0 \\
``body'' & 1 & 1.3 & 3.5 & 6 & 4.5 \\
``mane'' & - & - & - & - & 7 \\
\hline
\end{tabular}
}
\caption{Part lengths annotations for different animals obtained from ChatGPT in cm. Missing values are indicated by ``-''. Naming of animals taken from the Artemis~\cite{artemis} dataset.}
\label{tab:mapping_lengths}
\end{table}

%% file: author-kit/figures_supmat/length_anno.tex
\begin{figure}[t]
    \centering
    \includegraphics[width=0.95\linewidth]{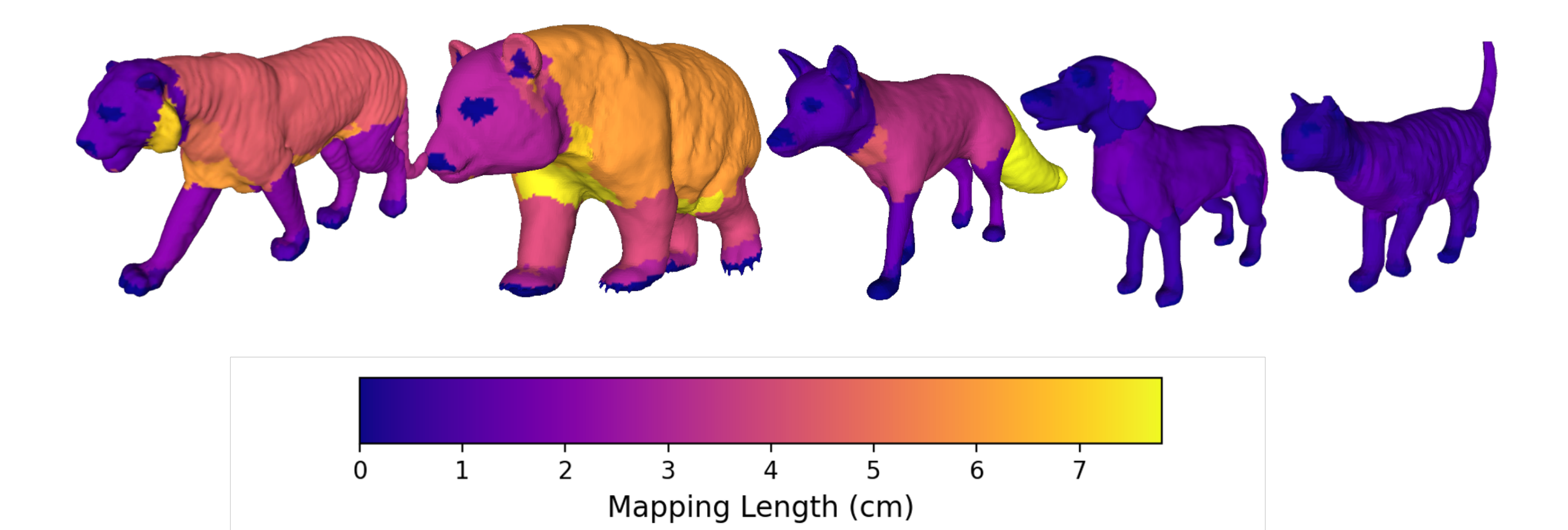}
    \caption{Length annotations obtained from ChatGPT for all animals.}
    \label{fig:length-annotation}
\end{figure}

%% file: author-kit/figures_supmat/effective_length.tex
\begin{figure}[t]
    \centering
    \includegraphics[width=0.95\linewidth]{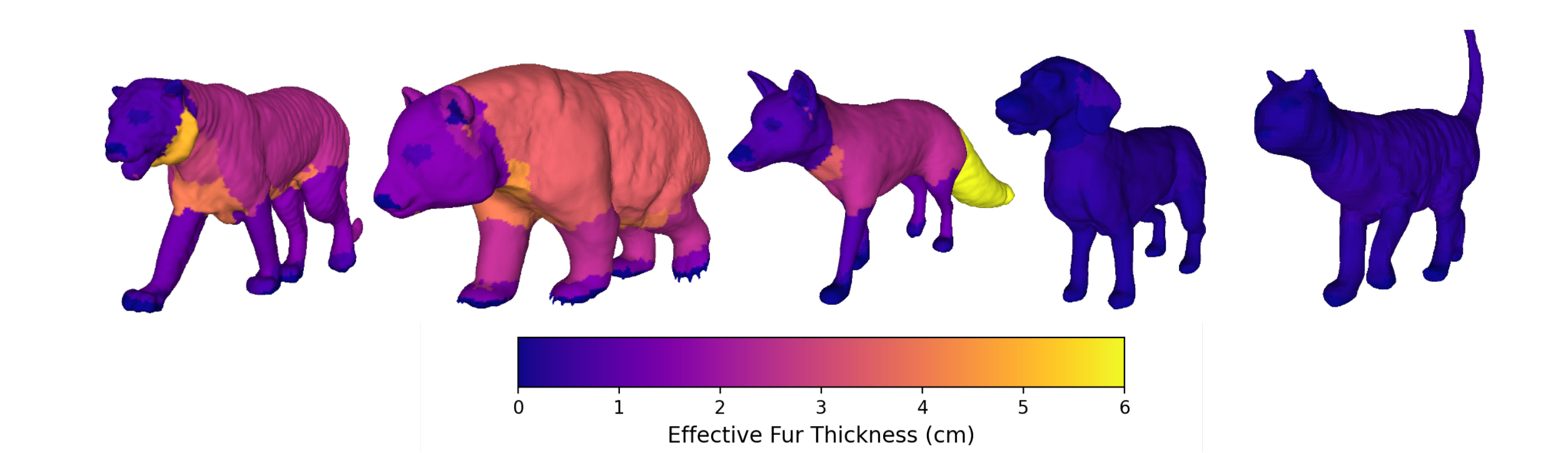}
    \caption{Effective fur thickness annotations obtained from ChatGPT for all animals.}
    \label{fig:effective-length-annotation}
\end{figure}

%% file: author-kit/figures_supmat/synth_data.tex
\begin{figure}[t]
    \centering
\includegraphics[width=0.98\linewidth,clip,trim=0cm 0cm 0cm 0cm]{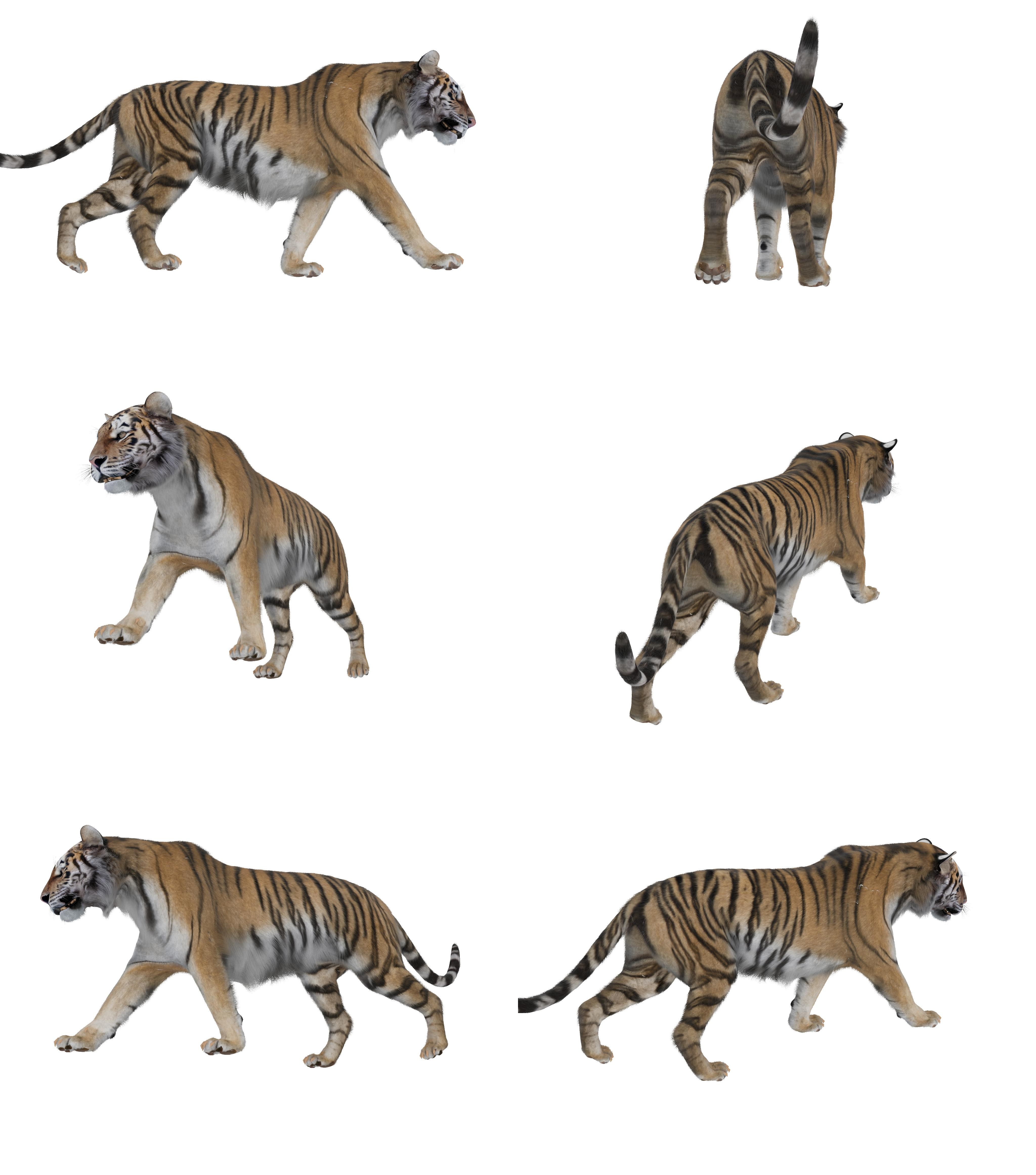}
\vspace{-0.5cm}
    \caption{Rendered in Blender~\cite{Blender} asset used for quantitative evaluation.}
    \label{fig:synth}
\end{figure}

%% file: author-kit/figures_supmat/editing_fur.tex
\begin{figure}[t]
    \centering
\includegraphics[width=0.9\linewidth,clip,trim=0cm 0cm 0cm 0cm]{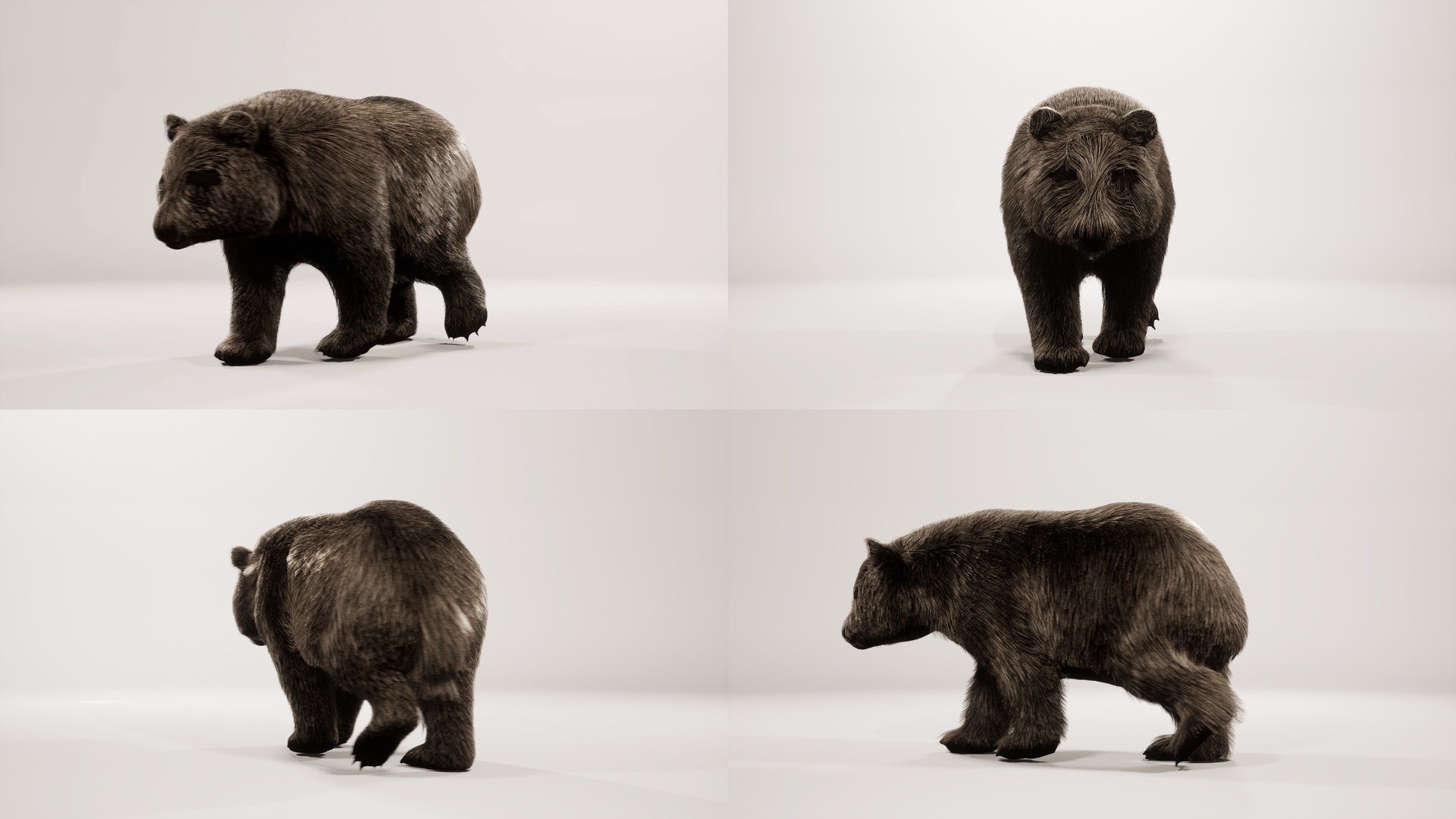}
    \caption{\textbf{Fur editing results.} Top-left: original image; top-right: face edited; bottom-left: tail edited; bottom-right: legs edited}
    \label{fig:editing_suppmat}
\end{figure}

%% file: author-kit/tables_suppmat/ablation_real_suppmat.tex
\definecolor{tabfirst}{rgb}{1, 0.7, 0.7} %
\definecolor{tabsecond}{rgb}{1, 0.85, 0.7} %
\definecolor{tabthird}{rgb}{1, 1, 0.7} %

\begin{table}[tbh]
\centering
\setlength{\tabcolsep}{4pt}  %
\renewcommand{\arraystretch}{1.2} %
\resizebox{\linewidth}{!}{
\begin{tabular}{l|cc|ccc|cc|c}
 & \multicolumn{8}{c}{\textit{``Cat''}} \\
\hline
Method & 
\multicolumn{2}{c|}{Fur length} & 
\multicolumn{3}{c|}{Fur curvature} &
\multicolumn{2}{c|}{Fur orientation} &
Fur silhouette \\
& $\mu_{L}$ $\pm$ $\sigma_L$ (cm)
& $\sigma_{\text{loc}}(L)$ $\downarrow$ 
& $\mathrm{Var}_{\text{glob}}(\kappa) \downarrow$
& $\mathrm{Var}_{\text{loc}}(\kappa) \downarrow$
& $\kappa_{\max} \downarrow$
& $\mathrm{Var}_{\text{loc}}(\text{dir}) \downarrow$
& $\mathrm{Var}_{\text{loc}}^{\text{first}}(\text{dir}) \downarrow$
& CD $\downarrow$\\
\hline
Ours                  &  \cellcolor{tabfirst}0.889 $\pm$ 0.29 & \cellcolor{tabsecond}0.046 &  \cellcolor{tabthird}0.0006 &  \cellcolor{tabfirst}0.000082 &  \cellcolor{tabfirst}1.70 &  \cellcolor{tabfirst}0.027 &  \cellcolor{tabthird}0.048 &  \cellcolor{tabfirst}0.000354 \\
GH (2nd stage)       &                      3.830 $\pm$ 2.77 &                      0.945 & \cellcolor{tabsecond}0.0005 &  \cellcolor{tabthird}0.000098 &                      2.79 &                      0.096 &                      0.093 &                      0.001907 \\
GH (3rd stage) &                      5.036 $\pm$ 2.59 &                      1.591 &                      0.0967 &                      0.151766 &                      3.13 &                      0.575 &                      0.632 &                      0.003013 \\
w/ opt. length             &                      8.255 $\pm$ 8.38 &  \cellcolor{tabthird}0.817 &                      0.0066 &                      0.000635 &                      3.04 &                      0.032 &  \cellcolor{tabfirst}0.044 &                      0.001143 \\
w/ fix. length            &  \cellcolor{tabthird}1.000 $\pm$ 0.00 &  \cellcolor{tabfirst}0.000 &                      0.0008 &  \cellcolor{tabthird}0.000098 & \cellcolor{tabsecond}2.06 &                      0.033 & \cellcolor{tabsecond}0.046 &                      0.000389 \\
w/ Unet           &  \cellcolor{tabfirst}0.889 $\pm$ 0.29 & \cellcolor{tabsecond}0.046 &  \cellcolor{tabfirst}0.0003 &                      0.000375 &                      2.68 &                      0.077 &                      0.119 &                      0.001778 \\
w/o defur              & \cellcolor{tabsecond}0.906 $\pm$ 0.29 & \cellcolor{tabsecond}0.046 &                      0.0010 &                      0.000313 &                      3.01 &                      0.046 &                      0.088 &                      0.001045 \\
w/o $\mathcal{L}_{\mathrm{chm}}$               &  \cellcolor{tabfirst}0.889 $\pm$ 0.29 & \cellcolor{tabsecond}0.046 &                      0.0007 & \cellcolor{tabsecond}0.000094 &  \cellcolor{tabthird}2.32 &  \cellcolor{tabthird}0.030 &                      0.056 &                      0.000445 \\
w/o $\mathcal{L}^{gpt}_{\text{dir}}$               &  \cellcolor{tabfirst}0.889 $\pm$ 0.29 & \cellcolor{tabsecond}0.046 &                      0.0010 &                      0.000218 &                      2.84 &                      0.033 & \cellcolor{tabsecond}0.046 & \cellcolor{tabsecond}0.000360 \\
w/o $\mathcal{L}_{\mathrm{shape}}$            &  \cellcolor{tabfirst}0.889 $\pm$ 0.29 & \cellcolor{tabsecond}0.046 &                      0.0034 &                      0.000448 &                      3.02 & \cellcolor{tabsecond}0.029 &                      0.085 &  \cellcolor{tabthird}0.000365 \\
\hline
\end{tabular}
}
\caption{Unsupervised geometry consistency metrics for length, direction, and curvature, evaluated for both local and global cases, as well as distance between roots and the outer surface. 
}
\label{tab:ablation_metrics_unsupervised_suppmat}
\end{table}

%% file: author-kit/figures_supmat/limitation_dog.tex
\begin{figure}[t]
    \centering
\includegraphics[width=0.8\linewidth,clip,trim=0cm 0cm 0cm 2.5cm]{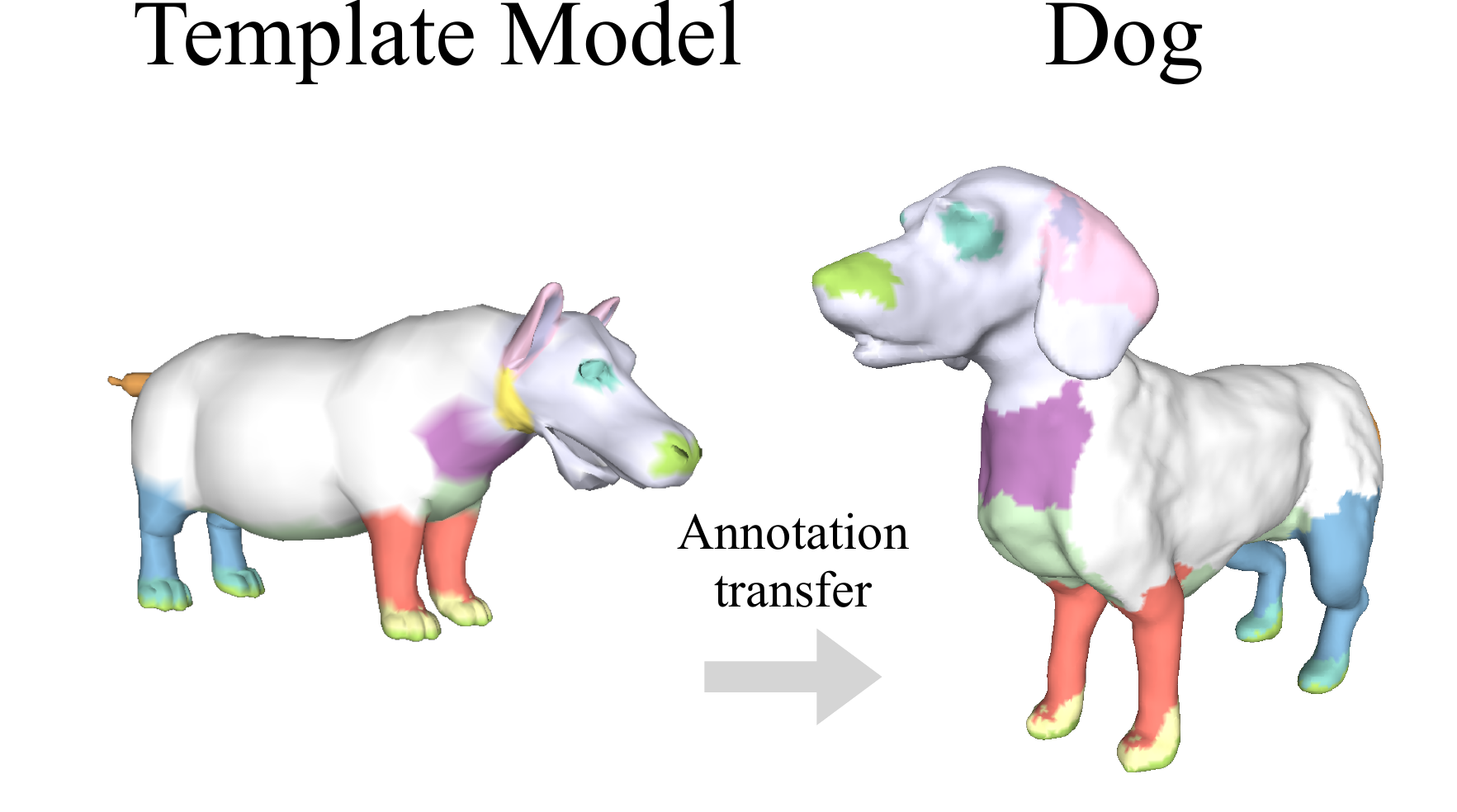}
    \caption{
    Transfer of the part annotations from the template (left) to a dog (right).
    Errors in body part transfer are mainly due to limitations of the fitted SMAL model, particularly for animals with large ears.}
    \label{fig:dog-limitation}
\end{figure}

%% file: author-kit/figures_supmat/ablations_num_views.tex
\begin{figure*}[t]
    \centering
    \includegraphics[width=\linewidth]{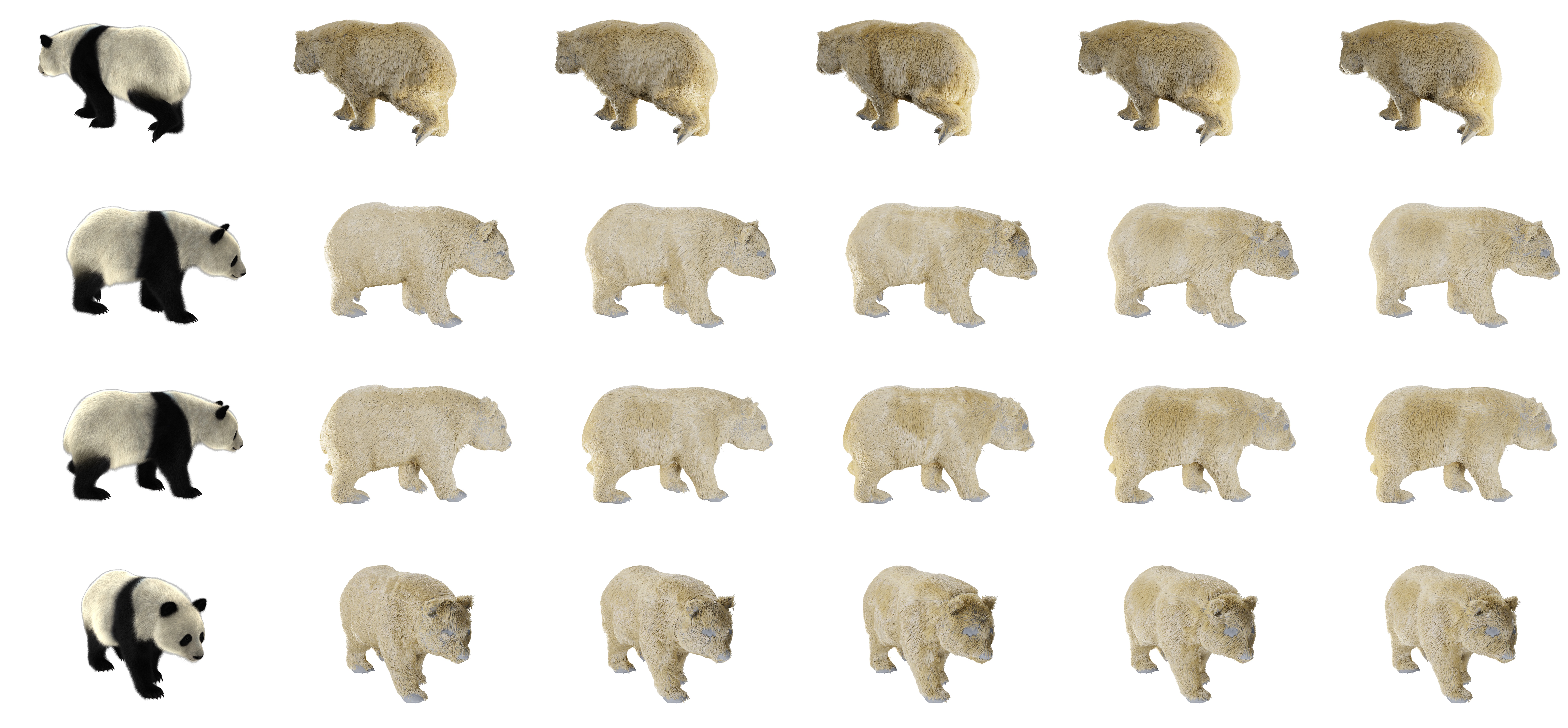}

    \makebox[0.162\linewidth]{Reference}%
    \makebox[0.162\linewidth]{Views = 1 }%
    \makebox[0.162\linewidth]{Views = 2}%
    \makebox[0.162\linewidth]{Views = 4}
    \makebox[0.162\linewidth]{Views = 8}%
    \makebox[0.162\linewidth]{Views = 16}%
    \caption{\textbf{Extended ablation study.} Qualitative evaluation of our method with different number of views. Even with a low number of views, our method is able to output reasonable fur reconstructions. However, the quality increases with more input views available.}
    \label{fig:ablation_results_supp_num_views}
\end{figure*}

%% file: author-kit/figures_supmat/ablations_supp.tex
\begin{figure*}[t]
    \centering
    \includegraphics[width=\linewidth]{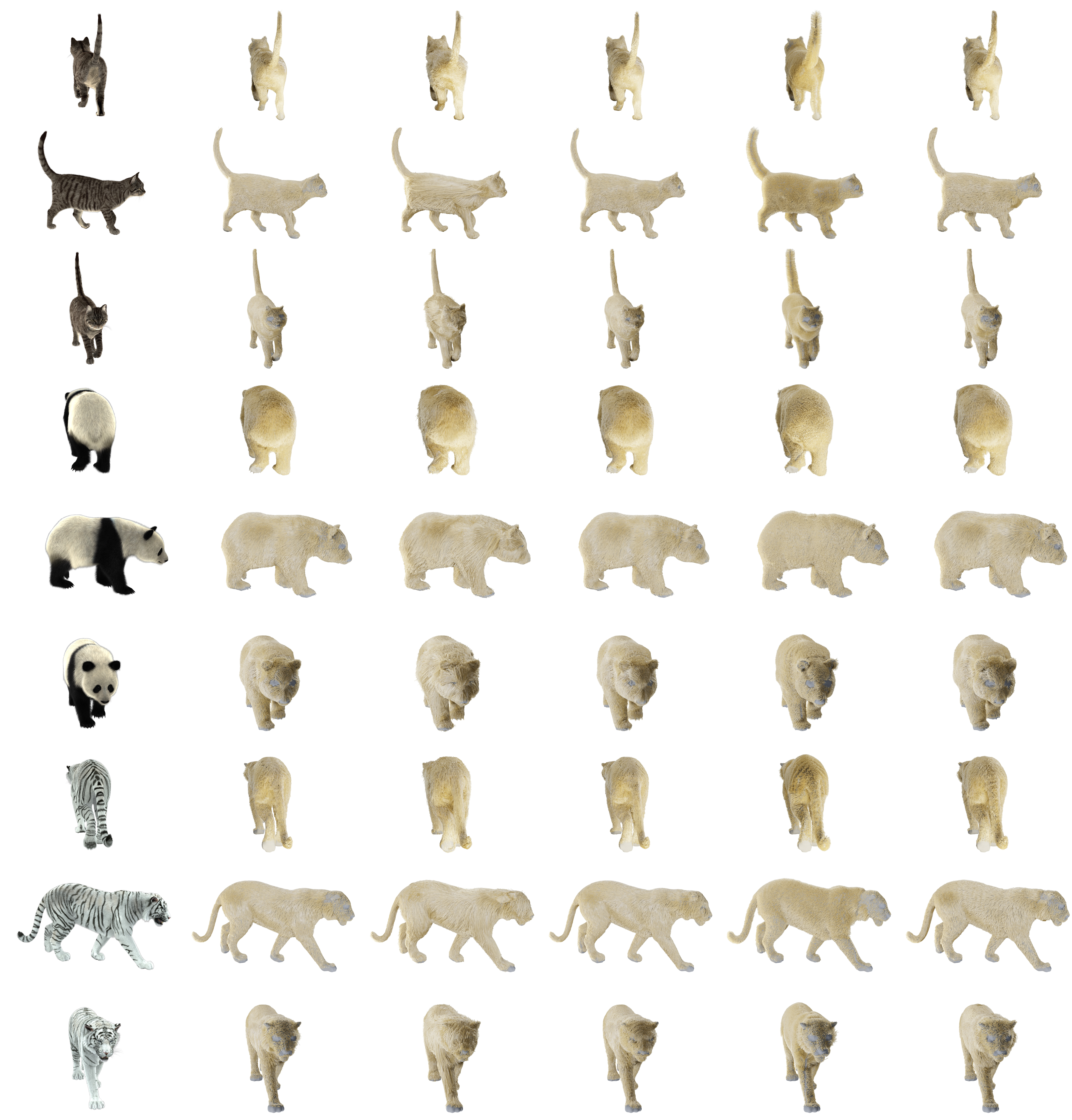}
    \vspace{0.25cm}
    \makebox[0.162\linewidth]{Reference}%
    \makebox[0.162\linewidth]{Ours}%
    \makebox[0.162\linewidth]{w/ opt. length}%
    \makebox[0.162\linewidth]{w/ fix. length}
    \makebox[0.162\linewidth]{ w/ UNet}%
    \makebox[0.162\linewidth]{w/o defur }%
    \caption{ \textbf{Extended ablation study.} Qualitative evaluation of our design choices regarding length, fur parametrization, and the importance of the defurring approach for accurate geometry modeling.}
    \label{fig:ablation_results_supp}
\end{figure*}

%% file: author-kit/figures_supmat/rendering.tex
\begin{figure*}[t]
    \centering
    \includegraphics[width=\linewidth]{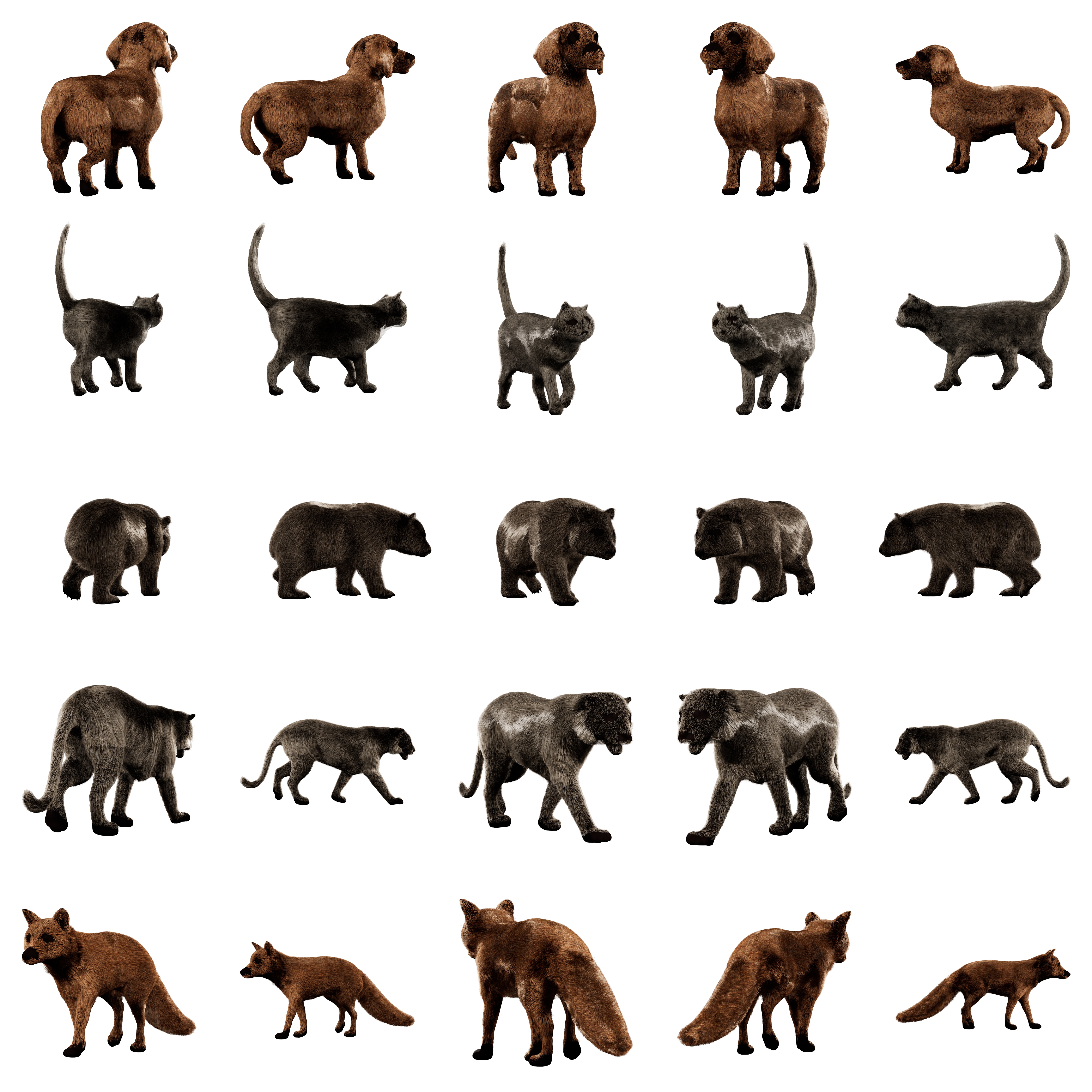}
    \caption{\textbf{Rendering examples.} Reconstructed strand-based fur could be easily rendered with predefined colors in Unreal Engine~\cite{unrealengine}.}
    \label{fig:unreal-rendering}
\end{figure*}

%% file: author-kit/figures_supmat/comparison_baselines_supp.tex
\begin{figure*}[t]
    \centering
    \includegraphics[width=\linewidth]{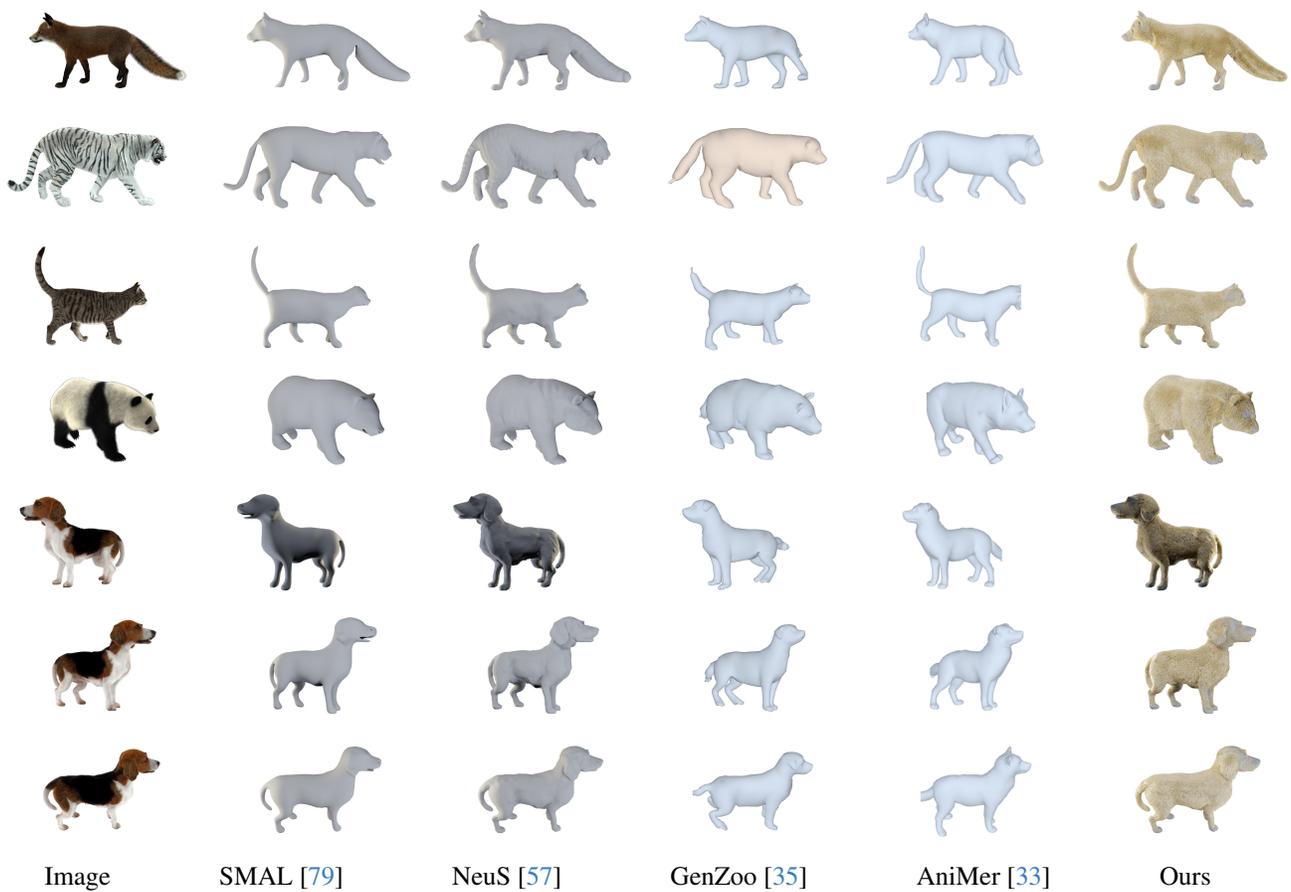}

    \begin{tabular*}{0.9\linewidth}{@{\extracolsep{\fill}}l l l l l l}
Image & SMAL~\cite{Zuffi:CVPR:2017} & NeuS~\cite{Wang2021NeuSLN} & GenZoo~\cite{niewiadomski2024generativezoo} & AniMer~\cite{lyu2025animer}  & Ours \\
\end{tabular*}
    
    \caption{\textbf{Extended comparison with baselines.} Surface reconstruction methods produce very coarse geometry.
    }
    
    \label{fig:comparison_suppmat_animals}
\end{figure*}

%% file: author-kit/figures_supmat/comparison_baselines_supp_GH.tex
\begin{figure*}[t]
    \centering
    \includegraphics[width=\linewidth]{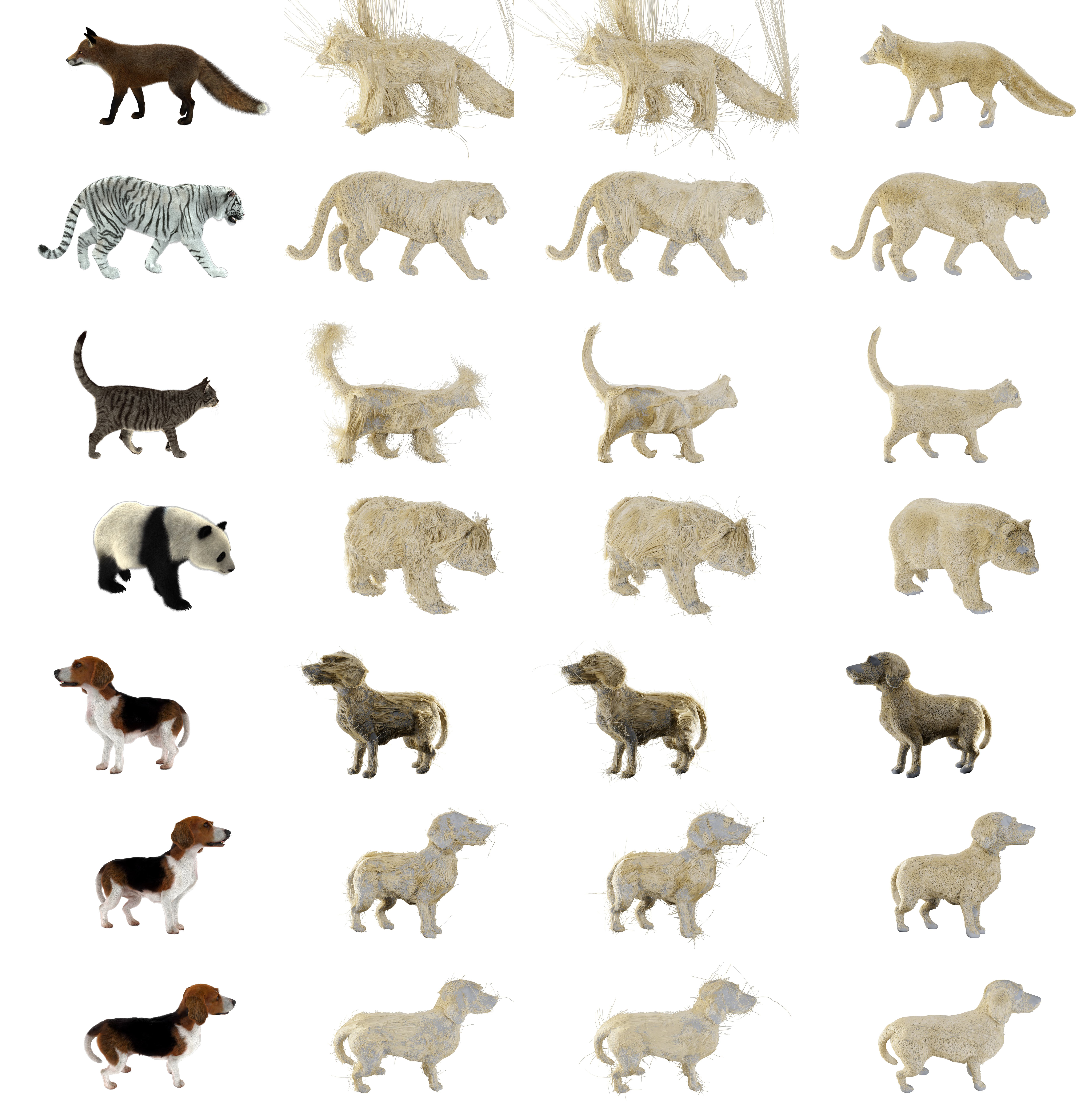}

    \begin{tabular*}{0.87\linewidth}{@{\extracolsep{\fill}}l l l l}
Image & GH (2nd stage)~\cite{zakharov2024gaussianhaircut} &GH (3rd stage)~\cite{zakharov2024gaussianhaircut} &  Ours \\
\end{tabular*}
    
    \caption{\textbf{Extended qualitative comparison with  Gaussian Haircut~\cite{zakharov2024gaussianhaircut}}, the state-of-the-art strand-based hair reconstruction method. Here, we compare with the results obtained after the second and third stages, see GH (2nd stage) and GH (3rd stage), respectively. Digital zoom-in is recommended.
    }
    
    \label{fig:comparison_suppmat_GH}
\end{figure*}